\documentclass{article} 
\usepackage{iclr2026_conference,times}

\usepackage{amsmath,amsfonts,bm}

\def\eqref#1{equation~\ref{#1}}

\def\1{\bm{1}}

\DeclareMathAlphabet{\mathsfit}{\encodingdefault}{\sfdefault}{m}{sl}
\SetMathAlphabet{\mathsfit}{bold}{\encodingdefault}{\sfdefault}{bx}{n}

\usepackage{hyperref}
\usepackage{url}

\usepackage{times}
\usepackage{latexsym}
\usepackage{fontawesome5}

\usepackage[T1]{fontenc}

\usepackage[utf8]{inputenc}

\usepackage{microtype}

\usepackage{inconsolata}

\usepackage{graphicx}
\usepackage{inconsolata}
\usepackage{makecell}
\usepackage{subfiles}

\usepackage[most]{tcolorbox} 

\usepackage{caption}
\usepackage{subcaption}
\usepackage{enumitem}
\usepackage{flushend}
\usepackage{balance}
\usepackage{lineno}
\usepackage{amsmath,amssymb,amsfonts}
\usepackage{algorithm}
\usepackage{algorithmic}
\usepackage{graphicx}
\usepackage{textcomp}
\usepackage{xcolor}
\usepackage{colortbl}
\usepackage{amsthm}
\usepackage{float}
\usepackage{multirow}
\usepackage{multicol}
\usepackage{color}
\usepackage{bm}
\usepackage{bbm}

\usepackage{pifont}
\usepackage{wrapfig}

\usepackage{booktabs}
\usepackage{multirow}
\usepackage{array}
\usepackage{tabularx}
\usepackage{fontawesome5}

\definecolor{ratingblue}{RGB}{220,230,250}     
\definecolor{metricgreen}{RGB}{225,245,230}    
\definecolor{rowgray}{gray}{0.97}              

\usepackage{booktabs}       
\usepackage{threeparttable} 
\newcommand{\greencheck}{\textcolor{green}{\ding{51}}}

\newcommand{\redcross}{\textcolor{red}{\ding{55}}}
\newcommand{\halfcheck}{\rlap{\textcolor{orange}{\ding{51}}} \hspace{-0.38em} {\textcolor{orange}{\ding{55}}}}
\definecolor{lightgray}{gray}{0.95}
\definecolor{lightblue}{RGB}{235,245,255}
\definecolor{headerblue}{RGB}{220, 235, 247}   
\definecolor{rowgray}{RGB}{245, 247, 250}       
\definecolor{accentblue}{RGB}{70, 130, 180}     
\definecolor{headergray}{RGB}{235,235,235}
\definecolor{subheadergray}{RGB}{245,245,245}
\definecolor{bookbg}{RGB}{248,244,252}
\definecolor{elecbg}{RGB}{240,247,242}
\definecolor{homebg}{RGB}{252,246,240}
\definecolor{carebg}{RGB}{245,242,250}
\definecolor{headerblue}{RGB}{220, 232, 245}     
\definecolor{subheader}{RGB}{237, 244, 251}    
\definecolor{rowlight}{RGB}{247, 251, 255}     
\definecolor{midrulecolor}{RGB}{180, 205, 230} 
\definecolor{headerblue}{RGB}{220, 232, 245}
\definecolor{subheader}{RGB}{237, 244, 251}
\definecolor{rowlight}{RGB}{247, 251, 255}
\definecolor{midrulecolor}{RGB}{180, 205, 230}
\definecolor{booksbg}{RGB}{245,248,255}
\definecolor{elecbg}{RGB}{246,252,248}
\definecolor{homebg}{RGB}{255,248,245}
\definecolor{carebg}{RGB}{248,246,255}

\definecolor{longbg}{RGB}{245,248,255}
\definecolor{memzerobg}{RGB}{246,252,248}
\definecolor{locomobg}{RGB}{255,248,245}
\definecolor{readbg}{RGB}{248,246,255}
\definecolor{bankbg}{RGB}{246,250,255}
\definecolor{amembg}{RGB}{255,250,245}

\definecolor{aspectcol}{RGB}{250,250,250}
\definecolor{rateblue}{RGB}{247,250,254}
\definecolor{rankblue}{RGB}{240,246,251}
\definecolor{summpink}{RGB}{253,248,251}
\definecolor{genpink}{RGB}{250,243,247}
\title{MemoryCD: Benchmarking Long-Context User Memory of 
LLM Agents for Lifelong Cross-Domain Personalization}

\author{
Weizhi Zhang$^{\blacksquare\spadesuit}$, \quad
Xiaokai Wei$^{\blacksquare}$,  \quad
Wei-Chieh Huang$^{\spadesuit}$,  \quad
Zheng Hui$^{\blacksquare\clubsuit}$,
\\ 
\quad \textbf{Chen Wang$^{\blacksquare}$,}
\quad \textbf{Michelle Gong$^{\blacksquare}$,} \quad
\textbf{Philip S. Yu}$^{\spadesuit}$\\
\quad $^{\blacksquare}$Roblox \quad
$^{\spadesuit}$University of Illinois Chicago \quad
$^{\clubsuit}$University of Cambridge \\
\quad \quad \quad \quad \quad \quad \quad \quad \quad 
\href{https://github.com/AgentMemoryWorld/MemoryCD}{\texttt{\textcolor{magenta}{github.com/AgentMemoryWorld/MemoryCD}}}
}

\iclrfinalcopy 
\begin{document}

\maketitle
\begin{abstract}
Recent advancements in Large Language Models (LLMs) have expanded context windows to million-token scales, yet benchmarks for evaluating memory remain limited to short-session synthetic dialogues. We introduce \textsc{MemoryCD}, the first large-scale, user-centric, cross-domain memory benchmark derived from lifelong real-world behaviors in the Amazon Review dataset. Unlike existing memory datasets that rely on scripted personas to generate synthetic user data, \textsc{MemoryCD} tracks authentic user interactions across years and multiple domains. We construct a multi-faceted long-context memory evaluation pipeline of 14 state-of-the-art LLM base models with 6 memory method baselines on 4 distinct personalization tasks over 12 diverse domains to evaluate an agent's ability to simulate real user behaviors in both single and cross-domain settings. Our analysis reveals that existing memory methods are far from user satisfaction in various domains, offering the first testbed for cross-domain life-long personalization evaluation. 
\end{abstract}

\section{Introduction}

\begin{wrapfigure}{r}{0.46\linewidth}
  \centering
  \vspace{-5pt}
  \includegraphics[width=\linewidth]{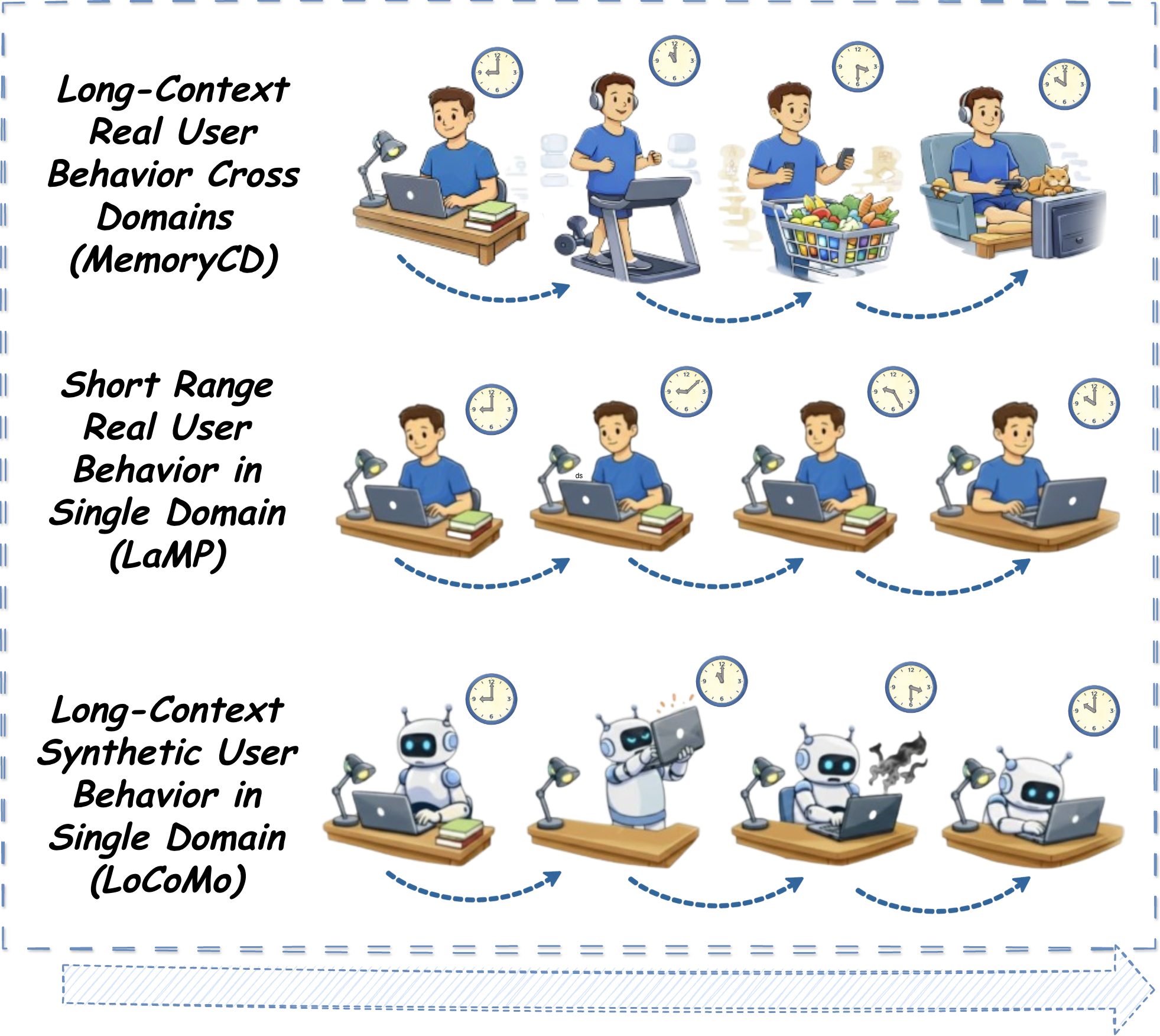}
  \vspace{-15pt}
  \caption{Comparison memory benchmarks: \textbf{\textsc{MemoryCD} (ours)} captures cross-domain real-user activities over long time horizons. \textbf{LaMP}~\citep{salemi2024lamp} focuses only on the short-term single-domain user behaviors; \textbf{LoCoMo}~\citep{maharana2024evaluating} represents non-authentic LLM-simulated user behaviors.}
  \label{fig:intro}
  \vspace{-8pt}
\end{wrapfigure}

The evolution of Large Language Models (LLMs) has ushered in a new era of general intelligence, with LLM-based agents demonstrating remarkable proficiency across standardized reasoning and generation tasks~\citep{naveed2025comprehensive, luo2025large, zhang2025web}. However, these advancements do not automatically translate into genuine user satisfaction~\citep{zhang2025personaagent}. 
For these agents to transition from generic tools to genuine personal assistants, they must possess the ability to retain, organize, and utilize the historical context of the user~\citep{cai2025large, chen2025persona}. 
Serving a real user requires more than responding to a single prompt; it requires grounding decisions in a coherent and persistent modeling of the user’s history, preferences, and evolving goals~\citep{tan2025prospect, zhang2025personaagent}. 
Memory provides a principled mechanism for enabling user-specific personalization in LLM agents~\citep{zhang2025survey}. As interactions accumulate over time, effective agents must consolidate long-horizon behavioral traces into abstract and durable user memories that support consistent personalization, improved task completion, and more natural user–agent interactions~\citep{sun2025scaling, yang2024swe, zhang2025personaagent}. This capability is critical for adapting to individual users, enabling agents to effectively support their work and daily life as lifelong lifelong personal assistant.

\begin{figure*}[ht]
  \centering
  \includegraphics[width=\linewidth]{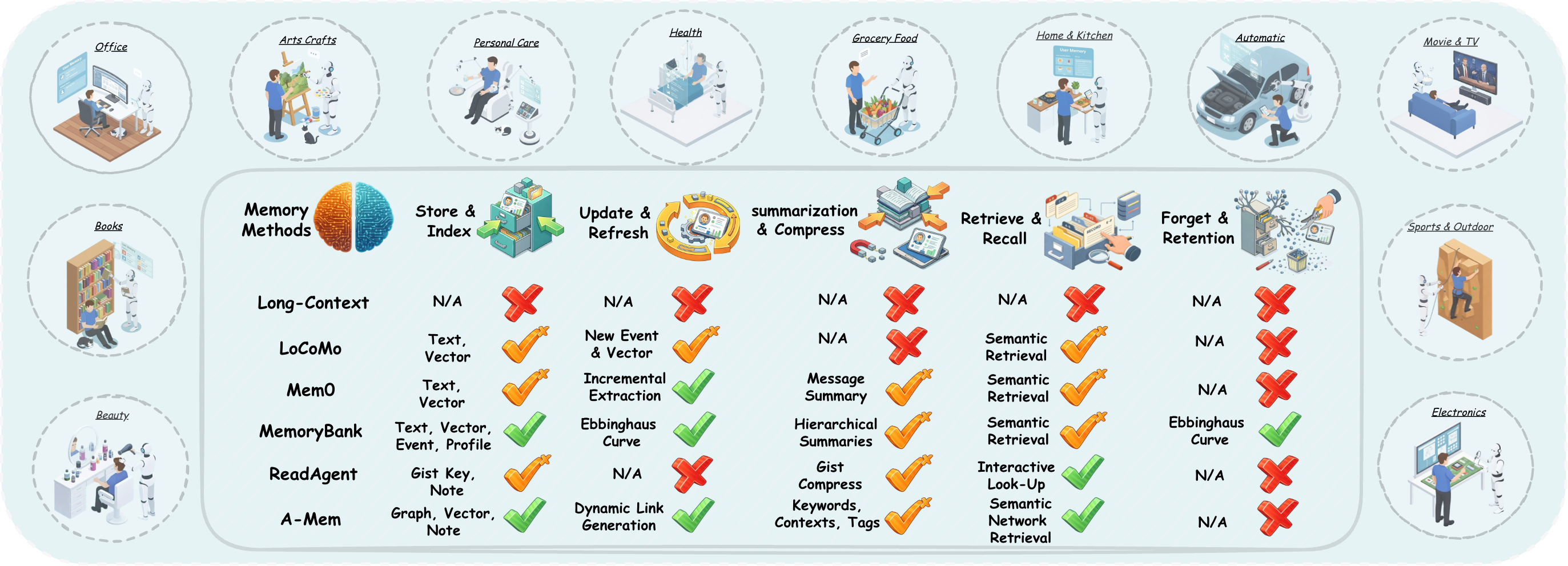}
  \caption{The \textsc{MemoryCD} benchmark spans \textbf{12 real-world domains} and evaluates \textbf{6 SOTA memory methods}. Different from other memory benchmarks targeting one specific memory stage (mostly retrieval), we design 4 basic tasks with 2 settings to provide end-to-end user satisfaction evaluation grounded on the lifelong real user behaviors (Table~\ref{tab:task-summary}).}
  \label{fig:main}
\vspace{-15pt}
\end{figure*}

Despite its importance, current memory benchmarks fall short of capturing user preferences and evaluating user satisfaction in real-world scenarios (as shown in Figure~\ref{fig:intro} and Table~\ref{tab: bench_compare}). First, the majority of memory benchmarks~\citep{maharana2024evaluating, ai2025memorybench, tan2025membench, zhao2025llms, wulong2025memeval, bei2026mem} are synthetic and constructed from LLM simulation that fail to provide user feedback from authentic users. Others with real user records, like LaMP~\citep{salemi2024lamp} and MSC~\citep{xu2022beyond}, are limited in length, offering only short sequences that do not stress the long-context windows available in modern LLMs. Additionally, user-centric datasets only evaluate preferences within isolated domains, ignoring the fact that user behavior is inherently across different domains. 
Real users express consistent personal tastes across various scenarios, and cross-domain memory can largely increase the cold-start user experiences~\citep{zhang2025cold, zhang2025llminit}.

From the evaluation perspective, existing memory benchmarks lack end-to-end, user-centric personalization assessments across diverse scenarios. As illustrated in Figure~\ref{fig:main}, memory systems~\citep{xu2025mem, zhong2024memorybank, zhu2025llm, zhang2026learning} generally involve five distinct stages; however, prevailing evaluation pipelines often treat memory as a narrow retrieval problem~\citep{maharana2024evaluating}. Tasks often reduce to detecting a specific fact in a sentence from a large context input. Therefore, they fail to evaluate the other stages of the memory lifecycle due to a lack of established ground-truth user feedback directly from the synthetic data.
In many production-level consumer applications, end-to-end evaluations like A/B testing~\citep{quin2024b} are preferred because they directly reflect user satisfaction. Real personalization requires agents to reason over long-term trajectories, infer latent preferences, and more importantly, act in ways aligned with user goals~\citep{zhang2025toward, shao2025dtrec}. Effective memory evaluation must therefore move beyond retrieval to user satisfaction assessments that examine whether stored and retrieved memory meaningfully influence downstream user decisions~\citep{huang2026rethinking}.

To address these limitations, we propose \textbf{\textsc{MemoryCD}}, the first benchmark that assesses long-context user-centric memory across domains with real user feedback. Based on data analysis on Amazon-Review~\citep{hou2024bridging}, we find that there is a large overlap in distinct domains, while they demonstrate the user preference covering many real-world applications, ranging from entertainment (\textit{Books} and \textit{Movies}) to healthcare (\textit{Health} and \textit{Personal Care}) and lifestyle (\textit{Sports} and \textit{Outdoors}). In these scenarios, users' preferences expressed in one domain and scenario influence their decisions in others, and an agent must systematically optimize the memory mechanisms to organize and utilize these user histories to provide coherent support across tasks. Furthermore, we implement 6 representative memory systems and find that existing memory mechanisms with long-context LLMs are still far from achieving user satisfaction. Nevertheless, our analysis shows that cross-domain user memory provides a promising pathway toward more comprehensive user understanding, although effective strategies for utilizing such memory remain largely unexplored.

In all, our work advances this direction through three key contributions: 
\begin{itemize}
\item We construct the first cross-domain long-context memory benchmark grounded in real user histories from the user-centric perspective for life-long personalization evaluation.

\begin{table*}[ht]
    \centering
    \setlength{\tabcolsep}{3pt}  
    \small
    \begin{tabular}{p{1.2cm} p{4.7cm} c c c c c c}
        \toprule
        \textbf{Data} &
        \textbf{Memory} &
        \textbf{Context} &
        \textbf{\# of } &
        \textbf{Long} &
        \textbf{User} &
        \textbf{Cross-} &
        \textbf{Real} \\ 
        \textbf{Source} &
        \textbf{Benchmarks} & 
        \textbf{Length} & 
        \textbf{Sessions} & 
        \textbf{Timespan} & 
        \textbf{Level} & 
        \textbf{Domain} &
        \textbf{Feedback} \\
        \midrule

        \multirow{8}{*}{Synthetic}
        & LoCoMo~\citep{maharana2024evaluating}   & 9K  & 35   & \redcross & \halfcheck & \redcross  & \redcross \\
        & PrefEval~\citep{zhao2025llms}    & 100K  & --   & \redcross  & \halfcheck & \halfcheck  & \redcross \\
        & LongMemEval~\citep{wulong2025memeval}   & 1.5M & 500 & \halfcheck & \halfcheck & \redcross  & \redcross \\ 
        & PersonaMem~\citep{jiang2025know}     & 1M & 60 & \greencheck & \halfcheck & \halfcheck & \redcross \\ 
        & MemBench~\citep{tan2025membench}    & 100K & 0.5K-8K & \halfcheck & \halfcheck & \redcross & \redcross \\ 
        & MemoryAgentBench~\citep{hu2025evaluating}  & 1M & -- & \halfcheck & \redcross & \redcross & \redcross \\ 
        & HaluMem~\citep{chen2025halumem}  & 1M & 120 & \greencheck & \greencheck & \redcross & \halfcheck \\    
        
        \midrule
        \multirow{3}{*}{Real User}
        & MSC~\citep{xu2022beyond} & 1K & 1-5   & \redcross  & \halfcheck & \redcross  & \halfcheck \\ 
        & LaMP~\citep{salemi2024lamp}  & 1K & -- & \halfcheck & \greencheck & \redcross  & \greencheck \\  
        & \textbf{\textsc{MemoryCD} (Ours)} 
                          & 400K & 1K+ & \greencheck & \greencheck & \greencheck & \greencheck \\
        \bottomrule
    \end{tabular}
    \caption{Comparison of representative memory benchmarks. \greencheck: fully covered; \halfcheck: partially covered; \redcross: not covered.
    \textbf{Synthetic} denotes simulated user behaviors.
    \textbf{Real User} denotes data constructed from real user interaction histories. 
    \textbf{\# of Sessions} is the number of sessions per individual users.}
    \label{tab: bench_compare}
\end{table*}

\item We design an end-to-end memory evaluation pipeline with 4 basic tasks in 2 memory settings to test whether agents can leverage memory for downstream user satisfaction, rather than merely long-context retrieval.

\item We conduct comprehensive evaluations with 14 frontier long-context LLMs and implemented 6 memory systems over 12 diverse domains, providing more insights and conclusions for future memory mechanism design.
\end{itemize}

\section{Task Settings and Problem Formulation}

We consider a memory evaluation setting for user-centric life-long personalization, in which the raw memory of each user $u \in U$ includes the interactions histories with various entities $i \in I$ across multiple years and domains $d \in D$. For each user, we construct a long-term behavioral record pool for user memory, defined as
\[
\mathcal{M}_u = \{(i, m_i, r_{u,i}, t_{u,i}, x_{u,i})\}_{i \in \mathcal{I}_u},
\]
where $i$ represents the item entity along with auxiliary metadata $m_i$, and $r_{u,i} \in \{1,\dots,5\}$ denotes the numerical preference rating. $t_{u,i}$ is the timestamp of the interaction, and $x_{u,i}$ contains the user textual preference opinion, including summarized title $x_{u,i}^{\prime}$ and full review text $x_{u,i}^*$. This sophisticated and real memory raw data represents the historical context of users over multiple years, during which an LLM agent must infer their long-term preference and linguistic tendencies to generalize to cross-domain behavioral patterns. Each task in \textsc{MemoryCD} is formalized as a mapping 
\[
f_\theta : (\mathcal{M}_u, \mathcal{X}) \rightarrow \mathcal{Y},
\]
where $\theta$ denotes the underlying LLM and memory mechanism, $\mathcal{X}$ specifies the task-dependent instruction input, and $\mathcal{Y}$ denotes the task-specific action output space. These tasks collectively evaluate whether the agent can approximate and align with the real user behavior by conditioning on users' prior behavior patterns in their memory. 

In following subsections, we first detail the four basic personalization tasks as in Table~\ref{tab:task-summary} and then clarify the two memory evaluation settings with single-domain and cross-domain memory sources.

\subsection{Personalized Rating Prediction}

The personalized rating prediction task measures an agent's ability to model quantitative user preferences. Given the user memory $\mathcal{M}_u$ and metadata $m_i$ associated with a target item, the objective is to predict the numerical preference rating $\hat{r}_{u,i}$ that user $u$ would assign to item entity $i$. Formally, the agent implements a regression function
\[
\hat{r}_{u,i} = f_\theta(\mathcal{M}_u, m_{i}),
\]
and performance is evaluated using deviation-based metrics: mean absolute error (MAE) and root mean-square error (RMSE)~\citep{hodson2022root}. This task probes the degree of long-term preference modeling and further calibration to numerical predictions.

\begin{table*}[ht]
\centering
\setlength{\tabcolsep}{5pt}
\small
\renewcommand{\arraystretch}{1.3}
\begin{tabular}{>{\columncolor{aspectcol}}p{1.2cm}
                >{\columncolor{rateblue}}p{2.8cm}
                >{\columncolor{rankblue}}p{2.8cm}
                >{\columncolor{summpink}}p{2.9cm}
                >{\columncolor{genpink}}p{2.8cm}}
\toprule
\cellcolor{aspectcol}\textbf{Aspect} &
\cellcolor{rateblue}\makecell{\ding{72}\ \quad \textbf{Rating} \\ \textbf{Prediction}} &
\cellcolor{rankblue}\makecell{\faListOl\ \quad \textbf{Item} \\ \textbf{Ranking}} &
\cellcolor{summpink}\makecell{\faPenFancy\ \quad \textbf{Review} \\ \textbf{Summarization}} &
\cellcolor{genpink}\makecell{\faAlignLeft\ \quad \textbf{Review} \\ \textbf{Generation}} \\
\midrule
\textbf{Category} &
User decision-making (quantitative preference modeling) &
User decision-making (relative preference modeling) &
User text generation (attitude \& sentiment reflection) &
User text generation (preference \& linguistic expression style) \\
\midrule
\textbf{Goal} &
Predict user's rating (1--5) for a target item. &
Rank candidate items by expected user interest. &
Generate a concise title reflecting user tone and sentiment. &
Generate full review text expressing user preferences and style. \\
\midrule
\textbf{Input} &
User memory $\mathcal{M}_u$; item metadata $m_i$ &
User memory $\mathcal{M}_u$; candidate set $C$ &
User memory $\mathcal{M}_u$; item metadata $m_i$; user review context $x_{u,i}^{*}$ &
User memory $\mathcal{M}_u$; item metadata $m_i$ \\
\midrule
\textbf{Output} &
Rating $\hat{r}_{u,i} \in \{1,\dots,5\}$ &
Ranking $\pi_u(C)$ &
Short title $x_{u,i}^{\prime}$ &
Full review $x_{u,i}^{*}$ \\
\midrule
\textbf{Metrics} &
MAE, RMSE &
NDCG@$K$ &
ROUGE-L, BLEU-1 &
ROUGE-L, BLEU-1 \\
\bottomrule
\end{tabular}
\caption{Summary of the four personalization tasks in \textsc{MemoryCD}.}
\label{tab:task-summary}
\end{table*}

\subsection{Personalized Item Ranking}

The personalized ranking task evaluates the agent's capacity to model relative preference strengths of the user. Given the memory $\mathcal{M}_u$ and a set of candidate items $C = \{i_1, \dots, i_k\}$ in the same domain, the agent predicts a ranking preference $\pi_u(C)$ reflecting the expected ordering of user interest:
\[
\pi_u(C) = f_\theta(\mathcal{M}_u, {m_{{i_1}}, \dots m_{{i_k}}}),
\]
Performance is assessed using NDCG@K~\citep{wang2013theoretical} (abbreviated as N@K), which quantify ranking consistency against the user's true preference signals. 

\subsection{Personalized Review Summarization}

The summarization task measures the agent's ability to produce concise, sentiment-aligned review titles grounded in the user's historical short review title patterns. Given the user memory $\mathcal{M}_u$, metadata $m_i$ for the target item, and the review full text $x_i^{\mathrm{*}}$, the model generates a short title
\[
{x}_{u,i}^{\prime} = f_\theta(\mathcal{M}_u, m_{i}, x_{u,i}^{*}).
\]
Evaluation is conducted using ROUGE-L~\citep{lin2004rouge} and BLEU-1 ~\citep{papineni2002bleu} to measure the alignment with the user’s tone, sentiment distribution, and lexical tendencies. 
This tests the capability to correctly integrate long-term preference into concise natural language outputs.

\subsection{Personalized Review Generation}
The personalized review generation task evaluates the agent’s ability to produce full-length review text that reflects both user's preference regarding item attributes and the user’s established expression style. Given the user memory $\mathcal{M}_u$ and the metadata $m_i$ of the target item, the agent generates a personalized review
\[
x_{u,i}^{*} = f_\theta(\mathcal{M}_u, m_i).
\]
Generated reviews are evaluated using ROUGE-L for sentence-level consistency and BLEU-1 for unigram-level lexical overlap, which collectively assess semantic fidelity, stylistic alignment, and preference-grounded expression. For simplicity, we refer to the two metrics ROUGE-L as ROUGE and BLEU-1 as BLEU in the following experimental tables. 

\subsection{Benchmark Memory Source Settings}

To evaluate how LLM agents utilize long-term user memory to align to the user preference, we design two complementary memory source settings that vary the relationship between the memory source and the target evaluations.

\paragraph{Single-domain memory.}
In the single-domain setting, the user memory $\mathcal{M}_u$ and the evaluation samples are drawn from the same domain. All historical interactions used for memory construction originate from the target domain. This setting evaluates an agent’s ability to perform within-domain long-context personalization of different scenarios.

\paragraph{Cross-domain memory.}
In the cross-domain setting, the agent is evaluated on a target domain while conditioning on user memory sourced from other domains. The memory formats are close to many real-world user histories, where $\mathcal{M}_u$ contains heterogeneous interactions that do not belong to the evaluation domain. This setting reflects realistic cold-start user~\citep{zhang2025cold} scenarios and tests whether memory mechanisms can transfer consistent and core preference signals of the user from different domains while suppressing irrelevant domain-specific noise.

\section{Dataset Analysis and Construction}
\label{sec:data}

\begin{wrapfigure}{r}{0.55\linewidth}
  \centering
  \vspace{-5pt}
  \includegraphics[width=\linewidth]{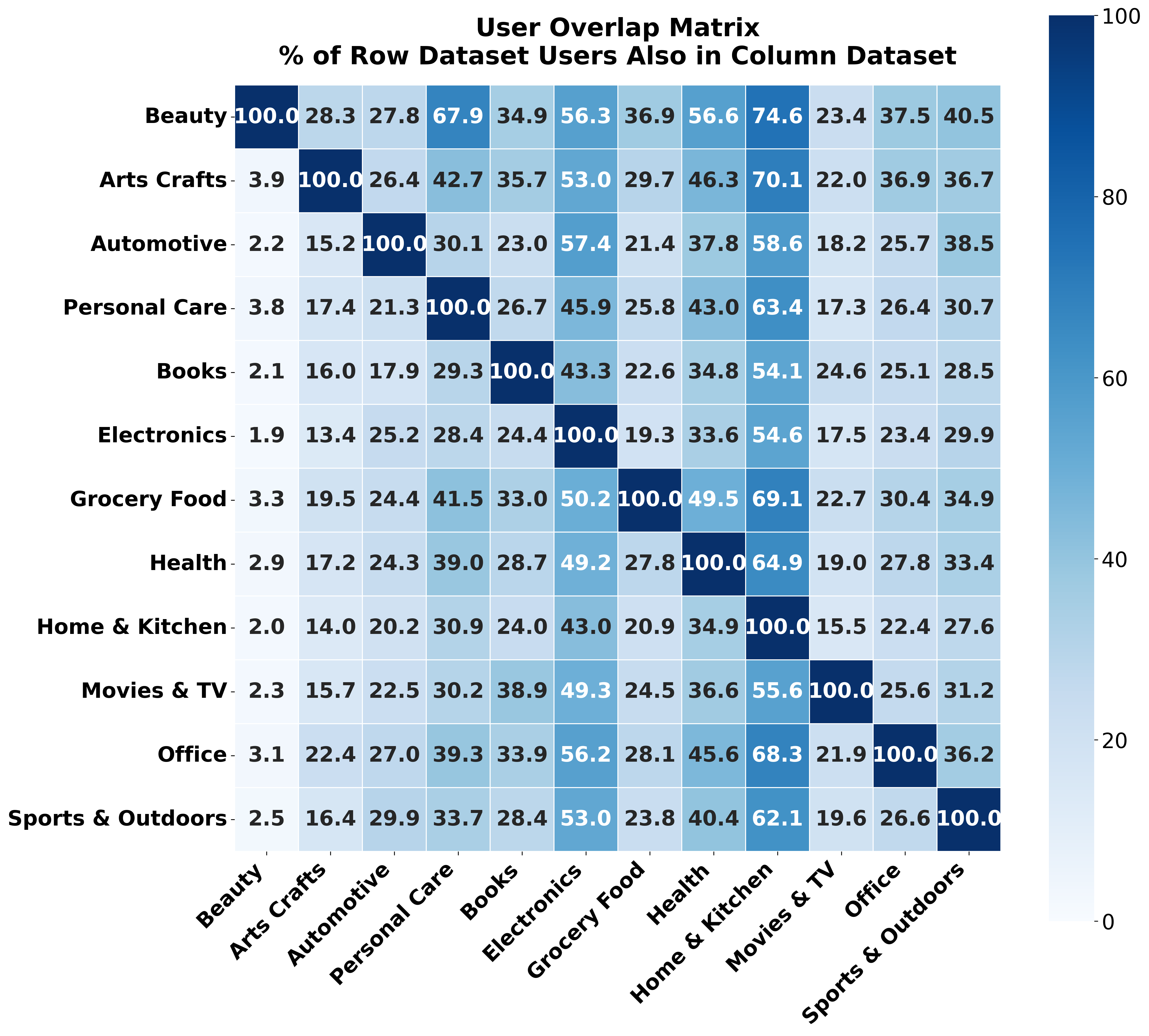}
  \vspace{-10pt}
  \caption{\textbf{Ratios of overlapping users across domains}. Each cell shows the percentage of users in one domain that also appear in another, revealing feasible cross-domain personalization settings.}
  \label{fig:my_figure}
  \vspace{-10pt}
\end{wrapfigure}

To curate the user-centric memory formulation, we need user-related datasets of long-context tokens with each user of long-range histories.
Here, \textsc{MemoryCD} is constructed from the Amazon Review~\citep{hou2024bridging} corpus and treats each user as a long-horizon behavioral process rather than a collection of isolated interactions.

To characterize the feasibility of cross-domain personalization, we analyze the degree of user overlap between domains. For each pair of domains $(d_i, d_j)$, we compute the ratio of users who appear in both domains relative to the total number of users in $d_i$. Figure~\ref{fig:my_figure} visualizes these overlap ratios as a heatmap. While many domain pairs exhibit limited overlap, we observe a non-trivial subset of users who are active across multiple domains, providing a natural foundation for cross-domain evaluation.

We construct \textsc{MemoryCD} across multiple domains and retain only users with sufficiently long interaction histories to ensure stable preference signals and long context memory lengths. Domain-specific minimum interaction thresholds are applied to filter sparse users, resulting in high-quality user memories with thousands of interactions per user in many domains. In the main experiments, we focus on four representative and complementary domains: {\textit{Personal Care}}, \textit{Books}, \textit{Electronics}, and \textit{Home \& Kitchen}. These domains exhibit diverse interaction patterns, textual styles, and large user overlaps, making them suitable for evaluating both within-domain and cross-domain user behaviors. More datasets processing details are attached in Appendix~\ref{sec:dataset_datails}.

\newcolumntype{C}[1]{>{\centering\arraybackslash}p{#1}}
\newcolumntype{L}[1]{>{\raggedright\arraybackslash}p{#1}}

\begin{table*}[!ht]
\centering
\footnotesize
\setlength{\tabcolsep}{1.2pt}
\small
\renewcommand{\arraystretch}{1.12}

\begin{tabularx}{\textwidth}{C{1.6cm} L{2.4cm} C{1.1cm} C{1.1cm} C{1.1cm} C{1.1cm} C{1.2cm} C{1.2cm} C{1.2cm} C{1.2cm}}
\toprule

\textbf{Dataset} & \textbf{Model} &
\multicolumn{2}{c}{\textbf{Rating} $\downarrow$} &
\multicolumn{2}{c}{\textbf{Ranking} $\uparrow$} &
\multicolumn{2}{c}{\textbf{Summarization} $\uparrow$} &
\multicolumn{2}{c}{\textbf{Generation} $\uparrow$} \\

& &
\textbf{MAE} & \textbf{RMSE} &
\textbf{N@1} & \textbf{N@3} &
\textbf{ROUGE} & \textbf{BLEU} &
\textbf{ROUGE} & \textbf{BLEU} \\
\midrule

\rowcolor{booksbg}
\cellcolor{white}{\multirow{7}{*}[-0.5ex]{\centering\textit{Books}}}
& GPT-5-Nano       & 0.543 & 0.872 & 0.135 & 0.164 & 0.122 & 0.055 & 0.110 & 0.145 \\
& GPT-5-Mini       & 0.464 & 0.816 & 0.161 & 0.190 & 0.138 & 0.063 & 0.130 & 0.174 \\
\rowcolor{booksbg}
\cellcolor{white}{}
& GPT-5            & \textbf{0.330} & \textbf{0.624} & 0.146 & 0.177 & 0.137 & 0.064 & 0.132 & 0.195 \\
& Claude-4 Opus    & 0.611 & 0.954 & 0.142 & 0.183 & 0.126 & 0.070 & 0.104 & 0.143 \\
\rowcolor{booksbg}
\cellcolor{white}{}
& Claude-4 Sonnet  & 0.670 & 1.000 & 0.109 & 0.163 & 0.145 & 0.089 & 0.110 & 0.151 \\
& Gemini-2.5 Flash & 0.412 & 0.750 & 0.127 & 0.178 & 0.182 & 0.092 & 0.142 & 0.159 \\
\rowcolor{booksbg}
\cellcolor{white}{}
& Gemini-2.5 Pro   & 0.401 & 0.762 & \textbf{0.206} & \textbf{0.229} & \textbf{0.208} & \textbf{0.128} & \textbf{0.162} & \textbf{0.207} \\
\midrule

\rowcolor{elecbg}
\cellcolor{white}{\multirow{7}{*}[-0.5ex]{\centering\textit{Electronics}}}
& GPT-5-Nano       & 0.435 & 0.789 & 0.252 & 0.357 & 0.173 & 0.076 & 0.108 & 0.124 \\
& GPT-5-Mini       & 0.354 & 0.701 & 0.366 & 0.488 & 0.202 & 0.099 & 0.132 & 0.158 \\
\rowcolor{elecbg}
\cellcolor{white}{}
& GPT-5            & 0.289 & 0.602 & \textbf{0.443} & \textbf{0.604} & 0.182 & 0.082 & 0.123 & 0.145 \\
& Claude-4 Opus    & \textbf{0.281} & \textbf{0.581} & 0.394 & 0.527 & {0.228} & 0.132 & 0.138 & 0.168 \\
\rowcolor{elecbg}
\cellcolor{white}{}
& Claude-4 Sonnet  & 0.374 & 0.749 & 0.354 & 0.511 & \textbf{0.228} & 0.139 & 0.144 & 0.176 \\
& Gemini-2.5 Flash & 0.358 & 0.744 & 0.309 & 0.447 & 0.211 & 0.104 & 0.146 & 0.173 \\
\rowcolor{elecbg}
\cellcolor{white}{}
& Gemini-2.5 Pro   & 0.337 & 0.719 & 0.378 & 0.561 & 0.222 & \textbf{0.141} & \textbf{0.161} & \textbf{0.200} \\
\midrule
\cmidrule(lr){2-10}

\rowcolor{homebg}
\cellcolor{white}{\multirow{7}{*}[-2.5ex]{\centering\parbox[c]{1.4cm}{\centering\textit{Home}\\ 0\\ \textit{\& Kitchen}}}}
& GPT-5-Nano       & 0.324 & 0.663 & 0.258 & 0.418 & 0.159 & 0.054 & 0.107 & 0.108 \\
& GPT-5-Mini       & 0.285 & 0.618 & 0.385 & 0.557 & 0.188 & 0.075 & 0.144 & 0.161 \\
\rowcolor{homebg}
\cellcolor{white}{}
& GPT-5            & \textbf{0.261} & \textbf{0.551} & \textbf{0.446} & \textbf{0.610} & 0.185 & 0.077 & 0.134 & 0.140 \\
& Claude-4 Opus    & 0.270 & 0.595 & 0.282 & 0.378 & 0.213 & 0.113 & 0.144 & 0.157 \\
\rowcolor{homebg}
\cellcolor{white}{}
& Claude-4 Sonnet  & 0.330 & 0.630 & 0.276 & 0.400 & 0.189 & 0.117 & 0.145 & 0.158 \\
& Gemini-2.5 Flash & 0.275 & 0.643 & 0.242 & 0.381 & 0.211 & 0.092 & 0.163 & 0.181 \\
\rowcolor{homebg}
\cellcolor{white}{}
& Gemini-2.5 Pro   & 0.303 & 0.679 & 0.315 & 0.444 & \textbf{0.222} & \textbf{0.133} & \textbf{0.178} & \textbf{0.207} \\
\midrule
\cmidrule(lr){2-10}

\rowcolor{carebg}
\cellcolor{white}{\multirow{7}{*}[-2.5ex]{\centering\parbox[c]{1.4cm}{\centering\textit{Personal}\\ 0\\ \textit{Care}}}}
& GPT-5-Nano       & 0.432 & 0.771 & 0.151 & 0.243 & 0.150 & 0.046 & 0.104 & 0.104 \\
& GPT-5-Mini       & 0.357 & 0.716 & 0.253 & 0.383 & 0.172 & 0.075 & 0.139 & 0.154 \\
\rowcolor{carebg}
\cellcolor{white}{}
& GPT-5            & \textbf{0.279} & \textbf{0.597} & \textbf{0.315} & \textbf{0.435} & 0.161 & 0.053 & 0.131 & 0.139 \\
& Claude-4 Opus    & 0.380 & 0.740 & 0.234 & 0.332 & 0.186 & 0.103 & 0.137 & 0.147 \\
\rowcolor{carebg}
\cellcolor{white}{}
& Claude-4 Sonnet  & 0.432 & 0.767 & 0.203 & 0.304 & 0.196 & \textbf{0.118} & 0.138 & 0.148 \\
& Gemini-2.5 Flash & 0.359 & 0.696 & 0.242 & 0.373 & 0.191 & 0.072 & 0.162 & 0.174 \\
\rowcolor{carebg}
\cellcolor{white}{}
& Gemini-2.5 Pro   & 0.323 & 0.665 & 0.266 & 0.417 & \textbf{0.200} & 0.110 & \textbf{0.183} & \textbf{0.204} \\
\bottomrule

\end{tabularx}

\caption{Performance comparison of frontier LLMs in single-domain with long-context memory prompting.}
\label{tab:all-tasks-combined}
\end{table*}




\section{Experiments}
\subsection{Experimental Setup}
\label{sec:setup}

We evaluate {14 frontier long-context LLM backbones} in combination with {6 state-of-the-art memory methods}. The model suite covers three most advanced LLM families, including the GPT series~\citep{achiam2023gpt}, Claude series~\citep{anthropic2024claude}, and Gemini series~\citep{comanici2025gemini}, spanning a wide range of context capacities and agentic behaviors. The memory methods include direct long-context prompting~\citep{liu2024lost} and five memory-augmented agents, including LoCoMo~\citep{maharana2024evaluating}, Mem0~\citep{chhikara2025mem0}, ReadAgent~\citep{lee2024human}, MemoryBank~\citep{zhong2024memorybank},  and A-Mem~\citep{xu2025mem},  that differ in how user information is stored, compressed, updated, retrieved, and forgotten. More memory methods descriptions are introduced in Appendix~\ref{app:baselines}.

For each user, we use the most recent three evaluation instances by default, resulting in {hundreds of test cases per domain}. Users are filtered using domain-specific minimum interaction thresholds (ranging from {50 to 1,000 historical interactions} as in Table~\ref{tab:domain_stats}) to ensure that each input memory constitutes a long-context, preference-revealing user history. 
To construct a challenging yet well-balanced cross-domain benchmark, we restrict the cross-domain dataset to {four domains}: \textit{Personal Care}, \textit{Books}, \textit{Electronics}, and \textit{Home \& Kitchen}. This selection explicitly balances the trade-off between the {number of eligible users} and the {length of interaction histories per user}, ensuring both sufficient population size and rich cross-domain memory.
Within single-domain or cross-domain settings, all models and memory methods are evaluated using the {same user memories}, task prompts, and decoding configurations to ensure fair and controlled comparison. Additional implementation and evaluation details are provided in Appendix~\ref{app:exp-details}. Note that, We do not adopt LLM-as-a-judge for text generation tasks due to its scalability limitations, inconsistency, and self-preference bias, as discussed in Appendix~\ref{app:judge}.

\begin{table*}[t]
\centering
\footnotesize
\setlength{\tabcolsep}{0.8pt}
\small
\renewcommand{\arraystretch}{1.15}

\begin{tabularx}{\textwidth}{C{1.9cm} L{2.2cm} C{1cm} C{1.1cm} C{1.1cm} C{1.1cm} C{1.5cm} C{1.2cm} C{1.5cm} C{1.2cm}}
\toprule
\textbf{Method} & \textbf{Model} &
\multicolumn{2}{c}{\textbf{Rating} $\downarrow$} &
\multicolumn{2}{c}{\textbf{Ranking} $\uparrow$} &
\multicolumn{2}{c}{\textbf{Summarization} $\uparrow$} &
\multicolumn{2}{c}{\textbf{Generation} $\uparrow$} \\

& &
\textbf{MAE} & \textbf{RMSE} &
\textbf{N@1} & \textbf{N@3} &
\textbf{ROUGE} & \textbf{BLEU} &
\textbf{ROUGE} & \textbf{BLEU} \\
\midrule

\rowcolor{longbg}
\cellcolor{white}{\multirow{3}{*}{\textbf{Long-Context}}}
& GPT-5
& \textbf{0.330} & \textbf{0.624}
& 0.146 & 0.177
& 0.137 & 0.064
& 0.132 & 0.195 \\

& Claude-4 Sonnet
& 0.670 & 1.000
& 0.109 & 0.163
& 0.145 & 0.089
& 0.110 & 0.151 \\

\rowcolor{longbg}
\cellcolor{white}{}
& Gemini-2.5 Pro
& 0.401 & 0.762
& \textbf{0.206} & \textbf{0.229}
& \textbf{0.208} & \textbf{0.128}
& \textbf{0.162} & \textbf{0.207} \\

\midrule

\rowcolor{memzerobg}
\cellcolor{white}{\multirow{3}{*}{\textbf{Mem0}}}
& GPT-5
& \textbf{0.419} & \textbf{0.698}
& \textbf{0.184} & 0.217
& 0.173 & 0.097
& 0.143 & \textbf{0.195} \\

& Claude-4 Sonnet
& 0.491 & 0.772
& 0.172 & 0.207
& 0.202 & 0.126
& 0.141 & 0.182 \\

\rowcolor{memzerobg}
\cellcolor{white}{}
& Gemini-2.5 Pro
& 0.464 & 0.803
& 0.180 & \textbf{0.221}
& \textbf{0.215} & \textbf{0.135}
& \textbf{0.162} & 0.185 \\

\midrule

\rowcolor{locomobg}
\cellcolor{white}{\multirow{3}{*}{\textbf{LoCoMo}}}
& GPT-5
& \textbf{0.360} & 0.643
& \textbf{0.232} & \textbf{0.253}
& 0.175 & 0.107
& 0.147 & \textbf{0.201} \\

& Claude-4 Sonnet
& 0.399 & 0.637
& 0.221 & 0.241
& 0.183 & 0.121
& 0.144 & 0.186 \\

\rowcolor{locomobg}
\cellcolor{white}{}
& Gemini-2.5 Pro
& 0.374 & \textbf{0.632}
& 0.230 & 0.248
& \textbf{0.192} & \textbf{0.127}
& \textbf{0.161} & 0.176 \\

\midrule

\rowcolor{readbg}
\cellcolor{white}{\multirow{3}{*}{\textbf{ReadAgent}}}
& GPT-5
& \textbf{0.431} & \textbf{0.700}
& \textbf{0.225} & \textbf{0.243}
& 0.174 & 0.095
& 0.142 & \textbf{0.193} \\

& Claude-4 Sonnet
& 0.476 & 0.727
& 0.202 & 0.238
& \textbf{0.188} & \textbf{0.120}
& 0.140 & 0.186 \\

\rowcolor{readbg}
\cellcolor{white}{}
& Gemini-2.5 Pro
& 0.479 & 0.807
& 0.210 & 0.235
& 0.181 & 0.117
& \textbf{0.155} & 0.173 \\

\midrule

\rowcolor{bankbg}
\cellcolor{white}{\multirow{3}{*}{\textbf{MemoryBank}}}
& GPT-5
& \textbf{0.356} & \textbf{0.627}
& 0.221 & 0.237
& 0.166 & 0.096
& \textbf{0.166} & \textbf{0.204} \\

& Claude-4 Sonnet
& 0.412 & 0.665
& 0.210 & 0.228
& 0.205 & \textbf{0.137}
& 0.145 & 0.188 \\

\rowcolor{bankbg}
\cellcolor{white}{}
& Gemini-2.5 Pro
& \textbf{0.378} & 0.679
& \textbf{0.228} & \textbf{0.242}
& \textbf{0.216} & 0.135
& 0.160 & 0.173 \\

\midrule

\rowcolor{amembg}
\cellcolor{white}{\multirow{3}{*}{\textbf{A-Mem}}}
& GPT-5
& \textbf{0.337} & \textbf{0.624}
& 0.210 & 0.232
& 0.162 & 0.091
& 0.143 & 0.203 \\

& Claude-4 Sonnet
& 0.371 & 0.633
& 0.150 & 0.188
& 0.212 & \textbf{0.150}
& 0.147 & 0.197 \\

\rowcolor{amembg}
\cellcolor{white}{}
& Gemini-2.5 Pro
& 0.390 & 0.687
& \textbf{0.217} & \textbf{0.237}
& \textbf{0.221} & 0.142
& \textbf{0.168} & \textbf{0.209} \\

\bottomrule
\end{tabularx}

\caption{Comparison of memory methods on \textit{Books} using representative LLMs.}
\label{tab:books_method_comparison}
\end{table*}

\subsection{Long-Context Prompting of LLMs in Single-Domain}
\label{sec:sd_long-context}
Although larger and more capable models generally perform better, the performance metrics are still far from perfect alignment with the user preference. In several cases, the gains of more advanced models are uneven across tasks. 
Table~\ref{tab:all-tasks-combined} reports the overall performance of state-of-the-art LLM base models across four personalization tasks in \textsc{MemoryCD} under long-context prompting, where each model directly consumes the full user interaction history without explicit memory abstraction or retrieval.
Improvements in rating or ranking scores do not translate into similar gains in personalized text generation quality. 
This decoupling suggests that personalization under long-context memory is not merely a naive problem naturally solved by the scaling law and motivates explicit and automatic memory mechanisms that can abstract, compress, and route user information more effectively than raw prompting.
Overall, GPT-5 and Gemini-2.5 Pro are the most competitive backbones across tasks, with GPT-5 excelling in rating and ranking tasks, and Gemini-2.5 Pro consistently leading in generation quality.
For example, GPT achieves the lowest rating prediction error in the \textit{Books} domain (RMSE 0.624) and delivers the best ranking performance in \textit{Electronics} (NDCG@3 0.604). In contrast, Gemini-2.5 Pro dominates language generation tasks, attaining the highest review generation scores across domains, including ROUGE-L of 0.162 in \textit{Books}.

\begin{figure*}[!ht]
    \centering
    \includegraphics[width=\linewidth]{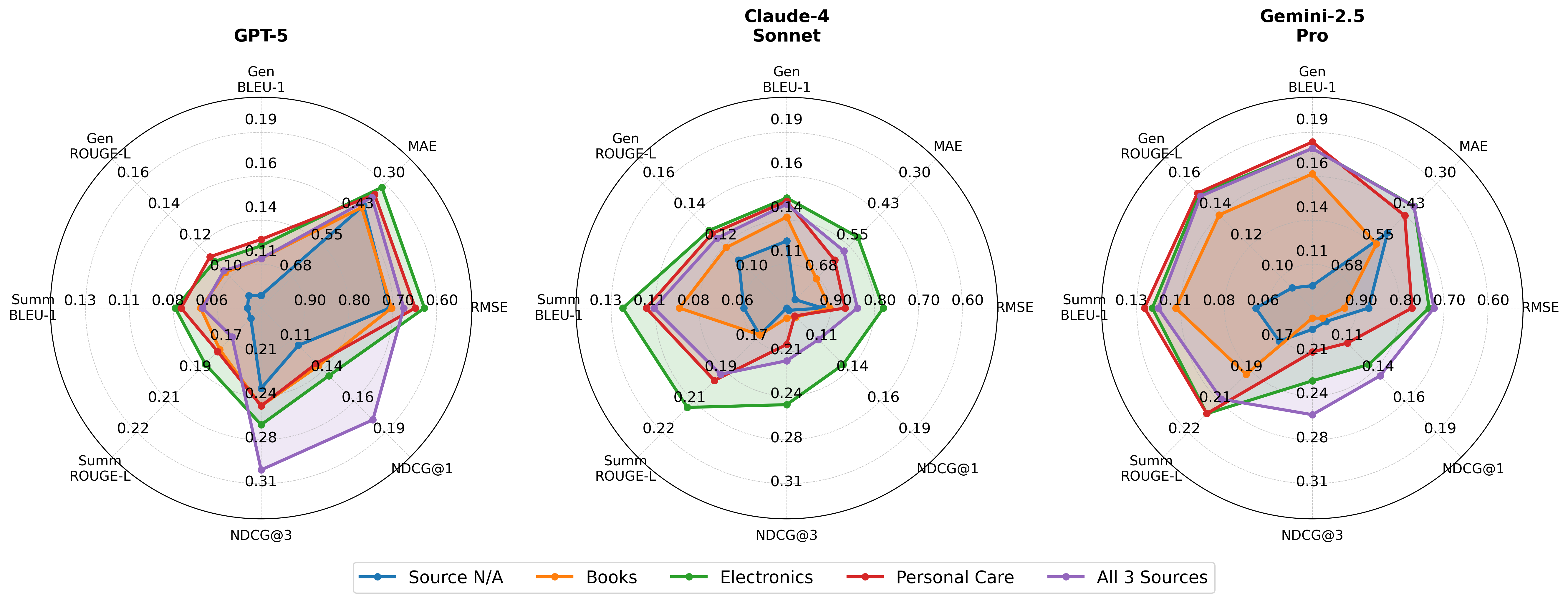}
    \caption{
    Performance comparisons using different memory sources evaluated on the \textit{Home \& Kitchen}.
    Each radar plot corresponds to a representative frontier long-context LLM ({GPT-5}, {Claude-4 Sonnet}, {Gemini-2.5 Pro}).
    Curves correspond to different memory sources: no memory ({Source N/A}),
    single-domain memories (\textit{Books}, \textit{Electronics}, \textit{Personal Care}),
    and aggregated cross-domain memory ({All 3 Sources}). Scales are unified for fair comparison (MAE and RMSE scales are reversed).
    }
    \vspace{20pt}
    \label{fig:MemoryCD-radar}
\end{figure*}

\begin{table*}[t]
\centering
\footnotesize
\setlength{\tabcolsep}{0.8pt}
\small
\renewcommand{\arraystretch}{1.15}

\begin{tabularx}{\textwidth}{C{1.9cm} L{2.2cm} C{1cm} C{1.1cm} C{1.1cm} C{1.1cm} C{1.5cm} C{1.2cm} C{1.5cm} C{1.2cm}}
\toprule
\textbf{Method} & \textbf{Model} &
\multicolumn{2}{c}{\textbf{Rating} $\downarrow$} &
\multicolumn{2}{c}{\textbf{Ranking} $\uparrow$} &
\multicolumn{2}{c}{\textbf{Summarization} $\uparrow$} &
\multicolumn{2}{c}{\textbf{Generation} $\uparrow$} \\

& &
\textbf{MAE} & \textbf{RMSE} &
\textbf{N@1} & \textbf{N@3} &
\textbf{ROUGE} & \textbf{BLEU} &
\textbf{ROUGE} & \textbf{BLEU} \\
\midrule

\rowcolor{longbg}
\cellcolor{white}{\multirow{3}{*}{\textbf{Long-Context}}}
& GPT-5
& \textbf{0.354} & \textbf{0.675}
& \textbf{0.179} & \textbf{0.299}
& 0.174 & 0.070
& 0.104 & 0.111 \\

& Claude-4 Sonnet
& 0.570 & 0.839
& 0.108 & 0.212
& 0.192 & 0.108
& 0.125 & 0.145 \\

\rowcolor{longbg}
\cellcolor{white}{}
& Gemini-2.5 Pro
& 0.391 & 0.723
& 0.140 & 0.255
& \textbf{0.204} & \textbf{0.119}
& \textbf{0.152} & \textbf{0.180} \\

\midrule

\rowcolor{memzerobg}
\cellcolor{white}{\multirow{3}{*}{\textbf{Mem0}}}
& GPT-5
& \textbf{0.367} & \textbf{0.661}
& \textbf{0.146} & \textbf{0.284}
& 0.189 & 0.094
& 0.112 & 0.122 \\

& Claude-4 Sonnet
& 0.596 & 0.823
& 0.140 & 0.255
& 0.197 & 0.116
& 0.127 & 0.146 \\

\rowcolor{memzerobg}
\cellcolor{white}{}
& Gemini-2.5 Pro
& 0.458 & 0.774
& 0.117 & 0.230
& \textbf{0.205} & \textbf{0.121}
& \textbf{0.149} & \textbf{0.176} \\

\midrule

\rowcolor{locomobg}
\cellcolor{white}{\multirow{3}{*}{\textbf{LoCoMo}}}
& GPT-5
& \textbf{0.348} & \textbf{0.653}
& \textbf{0.214} & \textbf{0.355}
& 0.196 & 0.094
& 0.130 & 0.130 \\

& Claude-4 Sonnet
& 0.522 & 0.776
& 0.155 & 0.272
& 0.189 & 0.106
& 0.130 & 0.150 \\

\rowcolor{locomobg}
\cellcolor{white}{}
& Gemini-2.5 Pro
& 0.402 & 0.759
& 0.205 & 0.349
& \textbf{0.202} & \textbf{0.119}
& \textbf{0.157} & \textbf{0.185} \\

\midrule

\rowcolor{readbg}
\cellcolor{white}{\multirow{3}{*}{\textbf{ReadAgent}}}
& GPT-5
& \textbf{0.408} & \textbf{0.690}
& \textbf{0.171} & \textbf{0.304}
& 0.188 & 0.088
& 0.114 & 0.126 \\

& Claude-4 Sonnet
& 0.590 & 0.815
& 0.142 & 0.244
& 0.183 & 0.102
& 0.127 & 0.148 \\

\rowcolor{readbg}
\cellcolor{white}{}
& Gemini-2.5 Pro
& 0.486 & 0.844
& 0.150 & 0.267
& \textbf{0.192} & \textbf{0.111}
& \textbf{0.152} & \textbf{0.180} \\

\midrule

\rowcolor{bankbg}
\cellcolor{white}{\multirow{3}{*}{\textbf{MemoryBank}}}
& GPT-5
& \textbf{0.326} & \textbf{0.634}
& \textbf{0.175} & \textbf{0.319}
& 0.206 & 0.109
& 0.124 & 0.139 \\

& Claude-4 Sonnet
& 0.507 & 0.758
& 0.133 & 0.263
& 0.201 & 0.126
& 0.137 & 0.159 \\

\rowcolor{bankbg}
\cellcolor{white}{}
& Gemini-2.5 Pro
& 0.397 & 0.726
& 0.134 & 0.260
& \textbf{0.214} & \textbf{0.134}
& \textbf{0.165} & \textbf{0.196} \\

\midrule

\rowcolor{amembg}
\cellcolor{white}{\multirow{3}{*}{\textbf{A-Mem}}}
& GPT-5
& \textbf{0.316} & \textbf{0.613}
& 0.132 & 0.239
& 0.194 & 0.092
& 0.113 & 0.122 \\

& Claude-4 Sonnet
& 0.514 & 0.766
& \textbf{0.146} & \textbf{0.274}
& 0.209 & 0.123
& 0.130 & 0.151 \\

\rowcolor{amembg}
\cellcolor{white}{}
& Gemini-2.5 Pro
& 0.363 & 0.695
& 0.117 & 0.230
& \textbf{0.221} & \textbf{0.134}
& \textbf{0.155} & \textbf{0.184} \\

\bottomrule
\end{tabularx}

\caption{Cross-domain evaluation of different memory methods on \textit{Home \& Kitchen}. Memory includes source domains of \textit{Books, Electronics, and Personal Care}.}
\label{tab:cd_method_comparison}
\end{table*}

\subsection{Memory Methods Evaluation in Single-Domain}
\label{sec:memory_baseline_results}

{Different memory methods show diverse advantages and improvements in different personalization tasks, but no one single method dominates}.
Table~\ref{tab:books_method_comparison} compares representative memory methods on \textit{Books} domain across four personalization tasks under three LLM backbones ({GPT-5}, {Claude-4 Sonnet}, {Gemini-2.5 Pro}). The evaluations on the \textit{Electronics}, \textit{Home \& Kitchen}, and \textit{Personal Care} are attached in Appendix Table~\ref{tab:sd_method_comparison_electronics}, Table~\ref{tab:sd_method_comparison_home_kitchen}, and Table~\ref{tab:sd_method_comparison_personal_care}, respectively.
Concretely, for rating prediction, long-context can be particularly strong (e.g., with GPT-5, long-context attains the best MAE of 0.330 among methods) because the relevant interactions are usually spread across the past user history. Inserting the full interaction history input can avoid the loss of extractions and allow strong backbones to directly aggregate weak but numerous cues. In contrast, more recent memory methods tend to help most when the task is {retrieval-sensitive} or {generation-sensitive}. For instance, the Mem0 pipeline based on the Claude-4 Sonnet boosts summarization ROUGE-L from 0.145 to 0.202 by removing redundant turns and gathering user details. LoCoMo enhances GPT-5's ranking by bringing in evidence that is important to preferences at the moment of decision. 
ReadAgent gist-memory compression with targeted lookup improves effective context usage and thus usually helps ranking compared to the raw long-context prompting (e.g., GPT-5 NDCG@1 from 0.146 to 0.225). In MemoryBank, the persistent user-profile updating can more directly stabilize personalized generation (e.g., GPT-5 generation ROUGE-L from 0.132 to 0.166) by maintaining an evolving user portrait rather than the static information. 
A-Mem’s structured note construction integrated with dynamic linking can especially benefit text metrics on some backbones (e.g., Claude-4 Sonnet BLEU-1 for summarization from 0.089 to 0.150) by enabling compositional retrieval over organized memories. However, although each mechanism yields different levels of improvements, no method is uniformly dominant across all four tasks and the overall metrics value are still low.

\subsection{Cross-Domain Memory Sources and LLMs Evaluation}
\label{sec:memory_baseline_results}

The effectiveness of cross-domain memory strongly depends on semantic alignment between source and target domains.
Figure~\ref{fig:MemoryCD-radar} illustrates that the usefulness of a memory source is highly domain-dependent. Across different backbones, \textit{Electronics} memories yield the strongest personalization on \textit{Home \& Kitchen}.  For GPT-5, MAE improves from  0.43 (Source N/A) to  0.30 with \textit{Electronics}; for Claude-4 Sonnet, MAE improves from 0.9 to 0.55.
This pattern is intuitive that \textit{Electronics} and \textit{Home \& Kitchen} share more overlapping product attributes and preference factors (e.g., brand, durability, functionality), making preference cues and evidence more reusable across domains. In contrast, {Books} memories of the same users show weaker transfer to the \textit{Home \& Kitchen} domain and many metrics are closely as those without additional memory sources, representing that the additional transferred memory is not always helpful but potentially introduces noise.

{Notably, aggregating cross-domain memories (\textbf{All 3 Sources}) often improves task performance.}
In all the backbones, \textbf{All 3 Sources} perform better than \textbf{Source N/A}. For GPT-5, \textbf{All 3 Sources} provides the best ranking performance, substantially outperforming \textbf{Source N/A}, indicating that broader personal information coverage helps retrieve relevant preference evidence when the target requires comparative reasoning. However, aggregation is not always optimal for error-based metrics. The best MAE/RMSE often comes from the most semantically aligned single source (notably \textit{Electronics}), while mixing in less-related domains can introduce distracting or conflicting information to hinder the prediction.

\subsection{Cross-Domain Memory Methods Evaluation}
\label{sec:memory_baseline_results}

Cross-domain personalization benefits from memory methods. In Table~\ref{tab:cd_method_comparison}, {Home \& Kitchen} with a memory pool aggregated from {Books/Electronics/Personal Care}, raw long-context prompting yields only moderate ranking quality, and memory methods mainly help by {filtering and compressing} heterogeneous histories into task-relevant cues, mitigating distraction from irrelevant evidence in long inputs. For ranking, LoCoMo long-horizon summarization shows the strongest transfer for GPT-5 and Gemini-2.5 Pro (NDCG@3 improves to 0.355 and 0.349, vs.\ 0.299 and 0.255). This improvement suggests that explicit memory routing is especially important when the model must suppress cross-domain noise. For the summarization and generation task, we observe A-Mem and MemoryBank perform the best, respectively. This shows that either the dynamic and consistent memory refreshment or removing noise can largely support cross-domain personalization on the generation task.

\section{Conclusion}

This paper introduces \textsc{MemoryCD}, a large-scale benchmark for evaluating ultra-long-context user memory in LLM agents using real-world, multi-domain interaction histories. By framing personalization as an end-to-end behavioral modeling problem, \textsc{MemoryCD} moves beyond retrieval-focused evaluation and assesses how memory usage impacts decision-making and personalized generation. Extensive experiments show that increasing model scale alone is insufficient for robust long-context personalization. Instead, memory design plays a critical role, with different mechanisms exhibiting distinct strengths across tasks. These findings underscore the need for principled memory abstractions that can effectively organize and exploit long-term user histories, and position \textsc{MemoryCD} as a foundation for future research on personalized and memory-driven LLM agents.

\section*{Limitations}
While \textsc{MemoryCD} provides a large-scale, real-user benchmark for long-context and cross-domain personalization, it has several limitations. First, although we include 12 domains and a carefully curated cross-domain subset, cross-domain evaluation is restricted to fixed memory operations rather than learned memory mechanisms~\citep{zhang2026memskill}, leaving broader domain transfer as future work. Second, our evaluation focuses on memory utilization and does not consider user privacy and safety guardrail~\citep{wu2025psg, huang2025deepresearchguard} over deployment time. Finally, large-scale training-based memory construction approaches~\citep{wang2025mem} are not incorporated into the benchmark, as most state-of-the-art long-context LLMs remain closed-source.

\bibliography{iclr2026_conference}
\bibliographystyle{iclr2026_conference}

\appendix

\clearpage

\section{Related Works}
\label{app: relates}

As summarized in Table~\ref{tab: bench_compare}, existing memory benchmarks for LLM agents differ substantially in data source, temporal coverage, personalization granularity, and evaluation scope. The majority of prior benchmarks are built on synthetic user behaviors, including LoCoMo~\citep{maharana2024evaluating}, PrefEval~\citep{zhao2025llms}, LongMemEval~\citep{wulong2025memeval}, PersonaMem~\citep{jiang2025know}, MemBench~\citep{tan2025membench}, and MemoryAgentBench~\citep{hu2025evaluating}. These benchmarks progressively extend context length and session counts, with some supporting million-token contexts or hundreds of simulated sessions. However, as shown in Table~\ref{tab: bench_compare}, most synthetic benchmarks only partially cover long-term temporal span and individual-level personalization, and they largely focus on single-domain settings without explicit user feedback. Consequently, evaluation often reduces memory to fact retrieval or isolated reasoning tasks, making it difficult to assess whether memory mechanisms genuinely support personalized decision-making or improve user satisfaction.

A smaller line of work evaluates memory using real user interaction histories, including MSC~\citep{xu2022beyond} and LaMP~\citep{salemi2024lamp}. These datasets introduce authentic user behaviors and, in the case of LaMP, individual-level preference modeling with real feedback signals. However, Table~\ref{tab: bench_compare} shows that existing real-user benchmarks are limited by short context lengths, few interaction sessions, and a lack of cross-domain coverage, which prevents stress-testing long-context memory and cross-domain preference transfer in modern LLMs. In contrast, \textsc{MemoryCD} uniquely integrates long-context inputs, long-term interaction histories, individual-level personalization, cross-domain behaviors, and explicit user feedback within a unified evaluation framework. This combination enables end-to-end assessment of whether memory mechanisms meaningfully influence downstream tasks across domains, addressing critical gaps left by prior synthetic and real-user benchmarks.

\section{Data Collection and Usage Compliance}

\textsc{MemoryCD} is constructed exclusively from the publicly released Amazon Reviews 2023 dataset~\citep{hou2024bridging}, following all data usage rules and licensing conditions specified by the dataset maintainers.\footnote{\url{https://amazon-reviews-2023.github.io/}} The dataset consists of anonymized, user-generated product reviews and ratings, collected and released for non-commercial academic research. No private, sensitive, or personally identifiable information is included or inferred at any stage of benchmark construction.

We strictly adhere to the original dataset schema and semantics. All user identifiers are treated as anonymous tokens, and \textsc{MemoryCD} does not attempt to de-anonymize users or link identities across external sources. Reviews, ratings, timestamps, and item metadata are used only as provided, without modification or augmentation beyond task-oriented filtering and aggregation.

\textsc{MemoryCD} is designed purely for evaluation purposes. The benchmark does not train or fine-tune language models on Amazon data, nor does it enable product-level inference or commercial recommendation use. All reported results measure an agent’s ability to condition on historical context of users at inference time, consistent with the intended academic use of the Amazon Reviews 2023 dataset.

\section{Datasets Details}
\label{sec:dataset_datails}

Table~\ref{tab:domain_stats} summarizes the domain-level statistics of the proposed ultra long-context, user-centric memory benchmark constructed from 12 product domains. The dataset exhibits extremely long user interaction histories, with context lengths ranging from 11K tokens (\textit{Beauty}) to over 314K tokens (\textit{Movies \& TV}). Highly active domains such as \textit{Books}, \textit{Electronics}, \textit{Personal Care}, and \textit{Movies \& TV} contain thousands of sessions per user (e.g., Books averages 1,552 sessions with a maximum of 3,888), reflecting rich long-term behavioral traces. To ensure meaningful personalization signals, each domain applies a minimum interaction threshold (50–1000 sessions depending on activity level), retaining only sufficiently active users for evaluation. Across domains, the benchmark around 100 users per domain, with each user contributing the most recent three samples for evaluation. The aggregated cross-domain setting further combines interactions across various scenarios, yielding the largest context length (387K tokens) and 5,862 sessions, covering 323 users and 969 samples, which provides a challenging testbed for evaluating memory mechanisms under ultra long-context and cross-domain personalization scenarios.

\begin{table*}[t]
\centering
\small
\setlength{\tabcolsep}{4pt}
\begin{tabular}{lrrrrrrr}
\toprule
\textbf{Domain} &
\makecell{\textbf{Context} \\ \textbf{Length}} &
\makecell{\textbf{Max \#} \\ \textbf{Sessions}} &
\makecell{\textbf{Avg \#} \\ \textbf{Sessions}} &
\makecell{\textbf{Min \#} \\ \textbf{Sessions}} &
\makecell{\textbf{Filtering} \\ \textbf{Thresholds}} &
\makecell{\textbf{\# of} \\ \textbf{Users}} &
\makecell{\textbf{\# of} \\ \textbf{Samples}} \\
\midrule
\rowcolor{gray!8}
Beauty              & 11{,}137   & 165   & 81    & 54    & 50    & 57  & 171 \\
Arts                & 49{,}101   & 801   & 338   & 204   & 200   & 50  & 150 \\
\rowcolor{gray!8}
Automotive          & 27{,}104   & 512   & 258   & 203   & 200   & 64  & 192 \\
Personal Care       & 202{,}096  & 2{,}973 & 839   & 504   & 500   & 128 & 384 \\
\rowcolor{gray!8}
Books               & 272{,}372  & 3{,}888 & 1{,}552 & 1{,}002 & 1{,}000 & 89  & 267 \\
Electronics         & 184{,}084  & 2{,}602 & 781   & 502   & 500   & 94  & 282 \\
\rowcolor{gray!8}
Grocery Food        & 124{,}204  & 1{,}524 & 760   & 510   & 500   & 45  & 135 \\
Health              & 180{,}757  & 2{,}640 & 831   & 501   & 500   & 75  & 225 \\
\rowcolor{gray!8}
Home \& Kitchen     & 175{,}039  & 2{,}754 & 768   & 501   & 500   & 110 & 330 \\
Movies \& TV        & 313{,}857  & 4{,}369 & 1{,}376 & 804   & 800   & 49  & 147 \\
\rowcolor{gray!8}
Office              & 71{,}497   & 921   & 331   & 203   & 200   & 25  & 75  \\
Sports \& Outdoors  & 64{,}527   & 858   & 296   & 203   & 200   & 66  & 198 \\
\midrule
\rowcolor{gray!8}
\textbf{Cross-Domain} & 386{,}736 & 5{,}862 & 973 & 243 & 200 & 323 & 969 \\
\bottomrule
\end{tabular}
\caption{
Domain-level statistics for the proposed ultra long-context user-centric memory benchmark.
Only users exceeding the minimum interaction threshold are retained; each user contributes
three evaluation samples.
}
\label{tab:domain_stats}
\end{table*}

\section{Memory Methods and Baselines}
\label{app:baselines}

We benchmark a diverse set of memory baselines that differ in how user information is stored, retrieved, and integrated into the model. These baselines span from naive full-history prompting to structured or compressed memory representations, allowing us to analyze how different memory designs trade off coverage, relevance, and robustness under ultra-long contexts.

\paragraph{Long-Context.}
The Long-Context~\citep{liu2024lost} method treats memory as a purely implicit mechanism by directly concatenating all historical user interactions into the prompt, relying on the expanded context window of modern LLMs rather than any explicit storage, indexing, or retrieval module. As shown in Figure~\ref{fig:main}, this approach lacks dedicated components for memory update, summarization, retrieval, or forgetting, and therefore serves as a minimal baseline that tests the raw capacity of long-context modeling. However, as the LLMs training paradigm naturally extend the context window size to the million-token level, this method serve as a simple yet strong baseline.

\paragraph{LoCoMo.}
LoCoMo~\citep{maharana2024evaluating} implements a retrieval-augmented generation (RAG) paradigm that couples explicit memory storage with vector indexing and on-the-fly retrieval to supplement the base LLM’s context window, addressing the limitations of purely concatenated long context. During inference, LoCoMo performs semantic retrieval to bring relevant memories into the prompt, improving response accuracy compared to the Long-Context baseline by reducing reliance on position biases and effectively foregrounding past events even when distant in the session. However, LoCoMo’s performance is still bounded by the quality of retrieved vectors and lacks sophisticated forget / retention mechanisms, which can lead to noise accumulation in long-running dialogues.

\paragraph{Mem0.}
Mem0~\citep{chhikara2025mem0} introduces a scalable, modular, and production-level long-term memory architecture that goes beyond simple RAG by dynamically extracting, consolidating, and maintaining salient information from past interactions to support persistent conversational memory across sessions. Rather than relying on concatenated context or raw chunk retrieval, Mem0 uses an LLM to identify key facts and preferences from recent dialogue and then stores these as structured memory entries.

\paragraph{ReadAgent.}
ReadAgent~\citep{lee2024human} is a human-inspired LLM agent designed to efficiently handle very long contexts by mimicking how people read and remember lengthy documents: it segments text into coherent “episodes,” compresses each segment into a gist memory, and selectively looks up detailed passages when needed to complete a task. Unlike simple RAG methods, ReadAgent explicitly performs store and index by grouping content into episodic memory and conducts summarization and compress via gist extraction, creating a compact representation of the original text. During inference, it uses an interactive look-up mechanism to retrieve relevant raw text based on the stored gists, enabling effective utilization of both global context and local details for comprehension tasks.

\paragraph{A-Mem.}
A-Mem~\citep{xu2025mem} introduces an agentic memory framework that dynamically organizes and evolves as knowledge-graph notes using principles inspired by the Zettelkasten method, enabling LLM agents to automatically generate structured notes with keywords, contextual descriptions, and semantic tags. When storing a new memory, A-Mem not only indexes it but also analyzes past memories to identify and instantiate dynamic links, thereby forming an interconnected knowledge network that continuously refines itself as new information arrives. During inference, this semantic network retrieval allows the agent to traverse linked memories and surface relevant historical information with greater contextual coherence than flat retrieval baselines.

\section{Experimental Details}
\label{app:exp-details}

\subsection{Evaluation Pipeline}
For each user $u$, we first construct the user memory $\mathcal{M}_u$ by aggregating all historical interactions. Across all domains, this process yields different long-context token length (details in Table~\ref{tab:domain_stats}). Memory-augmented methods are restricted to operating on $\mathcal{M}_u$ only, while all LLMs receive the same information through direct long-context prompting.

\subsection{Single-Domain and Cross-Domain Settings Datasets}
In the single-domain setting, both the user memory and evaluation queries are drawn from the same domain. We construct {12 single-domain benchmarks}, each containing around 100 users with long interaction histories, depending on the domain. In the cross-domain setting, we identify {323 users} who are simultaneously active in \textit{Books}, \textit{Electronics}, \textit{Home \& Kitchen}, and \textit{Personal Care}. For each such user, the memory aggregates interactions from \textit{Books}, \textit{Electronics}, and \textit{Personal Care} domains, while evaluating to the target domain \textit{Home \& Kitchen} by default.

\subsection{Implementation and Platforms}
All experiments are implemented in a unified evaluation framework. LLM backbones are accessed through official inference APIs, and all memory methods are implemented as modular components that interface with the same prompting templates and evaluation scripts. For each task, decoding hyperparameters are fixed across all models and methods. All evaluations are executed in a single run on the standardized cloud-based compute environment to avoid hardware-induced variability.

\subsection{Reproducibility and Fairness}
We fix random seeds for all stochastic components and evaluate each model and method configuration over the full set of test instances in each domain. No memory method has access to additional user profile, external corpora, or domain-specific information beyond the provided user memory. All metrics are computed based on identical implementations across settings to ensure that observed differences arise solely from model capacity and memory design rather than evaluation pipelines.

\section{Bias of LLM-as-a-judge for Frontier Models}\label{app:judge}

We focus on the two language-generation tasks: personalized review summarization and personalized review generation. The text outputs  are scored independently by three judge models of the frontier long-context LLMs: GPT-5, Claude 4 Sonnet, and Gemini-2.5 Pro. The judge reports scalar preference scores on a 1--10 scale upon comare the LLM results with the ground-truth real user drafted texts.

The results in Table~\ref{tab:llm_judge_bias_books} reveal two key patterns. First, individual LLM judges exhibit consistent self-preference bias, assigning higher scores to outputs generated by their own model. This leads to noticeable discrepancies in per-judge rankings despite small absolute score differences. Second, when aggregating scores across judges, the final ranking aligns with ROUGE-L and BLEU-1 based evaluation. In particular, Gemini-2.5 Pro ranks first in both tasks, while the relative ordering between GPT-5 and Claude 4 Sonnet varies by task. These findings suggest that although LLM-as-a-judge is biased at the individual level, simple aggregation can effectively mitigate this bias, albeit at the cost of increased computational overhead. Moreover, LLM-based evaluation is not fully stable or consistent: even under low-temperature settings, the same input can yield varying scores across runs. Therefore, we adopt ROUGE-L and BLEU-1 as scalable and reproducible metrics for evaluating text generation tasks.

\begin{table*}[t]
\centering
\small
\setlength{\tabcolsep}{5pt}
\begin{tabular}{llcccccc}
\toprule
\textbf{Task} & \textbf{Model} & \textbf{GPT-5} & \textbf{Claude} & \textbf{Gemini} & \textbf{Avg.} & \textbf{Final Rank} & \textbf{Our Rank} \\
\midrule
\rowcolor{gray!8}
\multirow{3}{*}{Review Summarization}
& GPT-5           & \textbf{5.9} & 5.3 & 5.1 & 5.43 & 3 & 3 \\
& Claude 4 Sonnet & 5.4 & \textbf{5.8} & 5.6 & 5.60 & 2 & 2 \\
\rowcolor{gray!8}
& Gemini-2.5 Pro  & 5.7 & 5.6 & \textbf{6.1} & 5.80 & 1 & 1 \\
\midrule
\rowcolor{gray!8}
\multirow{3}{*}{Review Generation}
& GPT-5           & \textbf{4.6} & 4.0 & 3.8 & 4.13 & 2 & 2 \\
& Claude 4 Sonnet & 4.0 & \textbf{4.3} & 4.0 & 4.10 & 3 & 3 \\
\rowcolor{gray!8}
& Gemini-2.5 Pro  & 4.5 & 4.3 & \textbf{4.8} & 4.53 & 1 & 1 \\
\bottomrule
\end{tabular}
\caption{LLM-as-a-judge evaluation with scores averaged across three judge models. Individual LLM judges exhibit self-preference bias, tending to favor outputs from their own model family. The final ranking is determined by the averaged scores, which align with our rank using ROUGE-L and BLEU-1 based evaluation in the Books domain.}
\label{tab:llm_judge_bias_books}
\end{table*}

\section{Evaluation of LLMs with long-context prompting in All Single Domains}
\label{app:llms_single_domain}

As presented in Table~\ref{tab:all-tasks-combined-8domains}, across the eight remaining single domains, we observe a clear {backbone specialization}: GPT-5 is strongest on decision tasks (rating and ranking), and Gemini-2.5 Pro is strongest on generated-related tasks (summarization/generation).
For rating prediction, GPT-5 achieves the best MAE in most of the domains, with particularly large margins in \textit{Arts \& Crafts}. For ranking, GPT-5 leads NDCG@3 in most of the domains (e.g., \textit{Automotive} 0.606; \textit{Toys \& Games} 0.557). In contrast, Gemini-2.5 Pro consistently leads {all domains} on summarization and generation. This trend shows a systematic advantage of Gemini-2.5 Pro in producing higher-overlap text outputs. 

In addition, achieving user satisfaction remains challenging and strongly task-dependent. For example, \textit{Beauty} consistently proves more difficult than structured product domains such as \textit{Arts \& Crafts}, suggesting that preference signals in certain domains are noisier or less directly predictive. Moreover, model rankings vary across task types. The backbone that performs best at predicting preferences (GPT-5) is not the one that best verbalizes preferences (Gemini-2.5 Pro), and the performance gap becomes particularly pronounced in generation-intensive domains. Overall, Table~7 reinforces that there is no universally best backbone across all tasks. Choosing the model and the memory method should be conditioned on whether the target objective is decision-centric (rating/ranking) or language-centric (summarization/generation).

\begin{table*}[!ht]
\centering
\footnotesize
\setlength{\tabcolsep}{0.8pt}  
\small
\renewcommand{\arraystretch}{1.15}

\begin{tabularx}{\textwidth}{C{1.6cm} L{2.5cm} C{1cm} C{1.1cm} C{1.1cm} C{1.1cm} C{1.5cm} C{1.2cm} C{1.5cm} C{1.2cm}}
\toprule
\textbf{Dataset} & \textbf{Model} &
\multicolumn{2}{c}{\textbf{Rating} $\downarrow$} &
\multicolumn{2}{c}{\textbf{Ranking} $\uparrow$} &
\multicolumn{2}{c}{\textbf{Summarization} $\uparrow$} &
\multicolumn{2}{c}{\textbf{Generation} $\uparrow$} \\
& &
\textbf{MAE} & \textbf{RMSE} &
\textbf{N@1} & \textbf{N@3} &
\textbf{ROUGE-L} & \textbf{BLEU-1} &
\textbf{ROUGE-L} & \textbf{BLEU-1} \\
\midrule

\multirow{3}{*}{\parbox[c]{1.4cm}{\centering\textbf{Beauty}}}
& GPT-5           & {0.544} & 0.946 & 0.175 & 0.290 & 0.203 & 0.085 & 0.231 & 0.118 \\
& Claude-4 Sonnet & \textbf{0.544} & \textbf{0.889} & \textbf{0.246} & 0.323 & 0.195 & 0.117 & 0.231 & 0.118 \\
& Gemini-2.5 Pro  & 0.719 & 1.221 & 0.228 & \textbf{0.349} & \textbf{0.230} & \textbf{0.129} & \textbf{0.284} & \textbf{0.159} \\
\midrule

\multirow{3}{*}{\parbox[c]{1.4cm}{\centering\textbf{Arts\\Crafts}}}
& GPT-5           & \textbf{0.167} & \textbf{0.440} & \textbf{0.367} & \textbf{0.531} & 0.209 & 0.092 & 0.246 & 0.131 \\
& Claude-4 Sonnet & 0.347 & 0.739 & 0.267 & 0.430 & 0.250 & 0.170 & 0.230 & 0.134 \\
& Gemini-2.5 Pro  & 0.327 & 0.787 & 0.327 & 0.486 & \textbf{0.292} & \textbf{0.201} & \textbf{0.307} & \textbf{0.182} \\
\midrule

\multirow{3}{*}{\textbf{Automotive}}
& GPT-5           & \textbf{0.318} & \textbf{0.851} & \textbf{0.443} & \textbf{0.606} & 0.226 & 0.117 & 0.159 & 0.078 \\
& Claude-4 Sonnet & 0.427 & 0.919 & 0.333 & 0.453 & 0.202 & 0.127 & 0.181 & 0.094 \\
& Gemini-2.5 Pro  & 0.427 & 1.021 & 0.328 & 0.477 & \textbf{0.291} & \textbf{0.188} & \textbf{0.228} & \textbf{0.118} \\
\midrule

\multirow{3}{*}{\parbox[c]{1.4cm}{\centering\textbf{Grocery\\Food}}}
& GPT-5           & \textbf{0.370} & \textbf{0.678} & \textbf{0.193} & \textbf{0.324} & 0.192 & 0.083 & 0.259 & 0.141 \\
& Claude-4 Sonnet & 0.578 & 0.919 & 0.148 & 0.256 & 0.151 & 0.101 & 0.197 & 0.116 \\
& Gemini-2.5 Pro  & 0.533 & 0.927 & 0.178 & 0.293 & \textbf{0.224} & \textbf{0.155} & \textbf{0.314} & \textbf{0.192} \\
\midrule

\multirow{3}{*}{\parbox[c]{1.4cm}{\centering\textbf{Health}}}
& GPT-5           & \textbf{0.382} & \textbf{0.712} & \textbf{0.302} & \textbf{0.428} & 0.149 & 0.060 & 0.250 & 0.132 \\
& Claude-4 Sonnet & 0.627 & 0.955 & 0.196 & 0.324 & 0.141 & 0.089 & 0.207 & 0.118 \\
& Gemini-2.5 Pro  & 0.431 & 0.803 & 0.173 & 0.322 & \textbf{0.180} & \textbf{0.106} & \textbf{0.310} & \textbf{0.188} \\
\midrule

\multirow{3}{*}{\parbox[c]{1.4cm}{\centering\textbf{Movies\\\& TV}}}
& GPT-5           & \textbf{0.524} & \textbf{0.869} & \textbf{0.367} & \textbf{0.474} & 0.131 & 0.063 & 0.243 & 0.142 \\
& Claude-4 Sonnet & 0.667 & 1.030 & 0.252 & 0.350 & 0.129 & 0.087 & 0.205 & 0.119 \\
& Gemini-2.5 Pro  & 0.612 & 1.010 & 0.265 & 0.391 & \textbf{0.152} & \textbf{0.105} & \textbf{0.287} & \textbf{0.170} \\
\midrule

\multirow{3}{*}{\parbox[c]{1.4cm}{\centering\textbf{Office}}}
& GPT-5           & \textbf{0.427} & 0.980 & \textbf{0.320} & \textbf{0.498} & 0.221 & 0.097 & 0.278 & 0.158 \\
& Claude-4 Sonnet & 0.533 & 0.980 & 0.227 & 0.351 & 0.181 & 0.113 & 0.273 & 0.151 \\
& Gemini-2.5 Pro  & 0.467 & \textbf{0.917} & {0.320} & 0.441 & \textbf{0.252} & \textbf{0.169} & \textbf{0.341} & \textbf{0.206} \\
\midrule

\multirow{3}{*}{\parbox[c]{1.4cm}{\centering\textbf{Toys\\\& Games}}}
& GPT-5           & \textbf{0.266} & \textbf{0.556} & \textbf{0.416} & \textbf{0.557} & 0.199 & 0.103 & 0.257 & 0.142 \\
& Claude-4 Sonnet & 0.519 & 0.892 & 0.385 & 0.508 & 0.229 & 0.149 & 0.206 & 0.119 \\
& Gemini-2.5 Pro  & 0.403 & 0.809 & 0.383 & 0.547 & \textbf{0.247} & \textbf{0.164} & \textbf{0.323} & \textbf{0.203} \\
\bottomrule
\end{tabularx}

\caption{Overall performance of representative LLMs across four tasks in the rest eight single domains of \textsc{MemoryCD} with long-context memory prompting.}
\label{tab:all-tasks-combined-8domains}
\end{table*}

\section{Evaluation of All LLMs with Representative Memory Methods in Single Domains Setting}
\label{app:llms_single_domain}

\begin{figure*}[!ht]
    \centering
    \includegraphics[width=\linewidth]{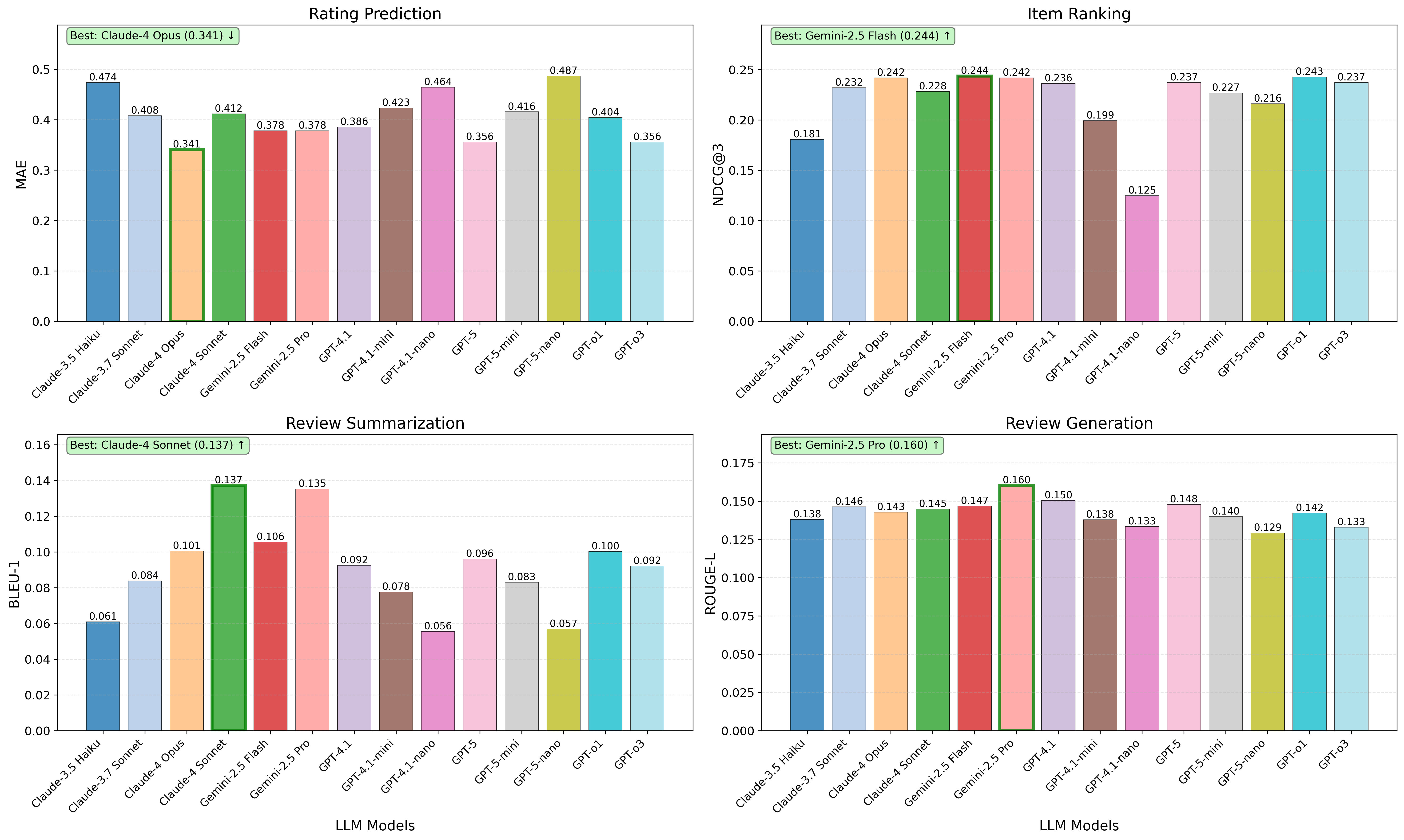}
    \caption{
    Performance comparisons of 14 frontier long-context LLMs on Books based on MemoryBank.
    }
    \label{fig:sd_memorybank_books}
\end{figure*}

\begin{figure*}[!ht]
    \centering
    \includegraphics[width=\linewidth]{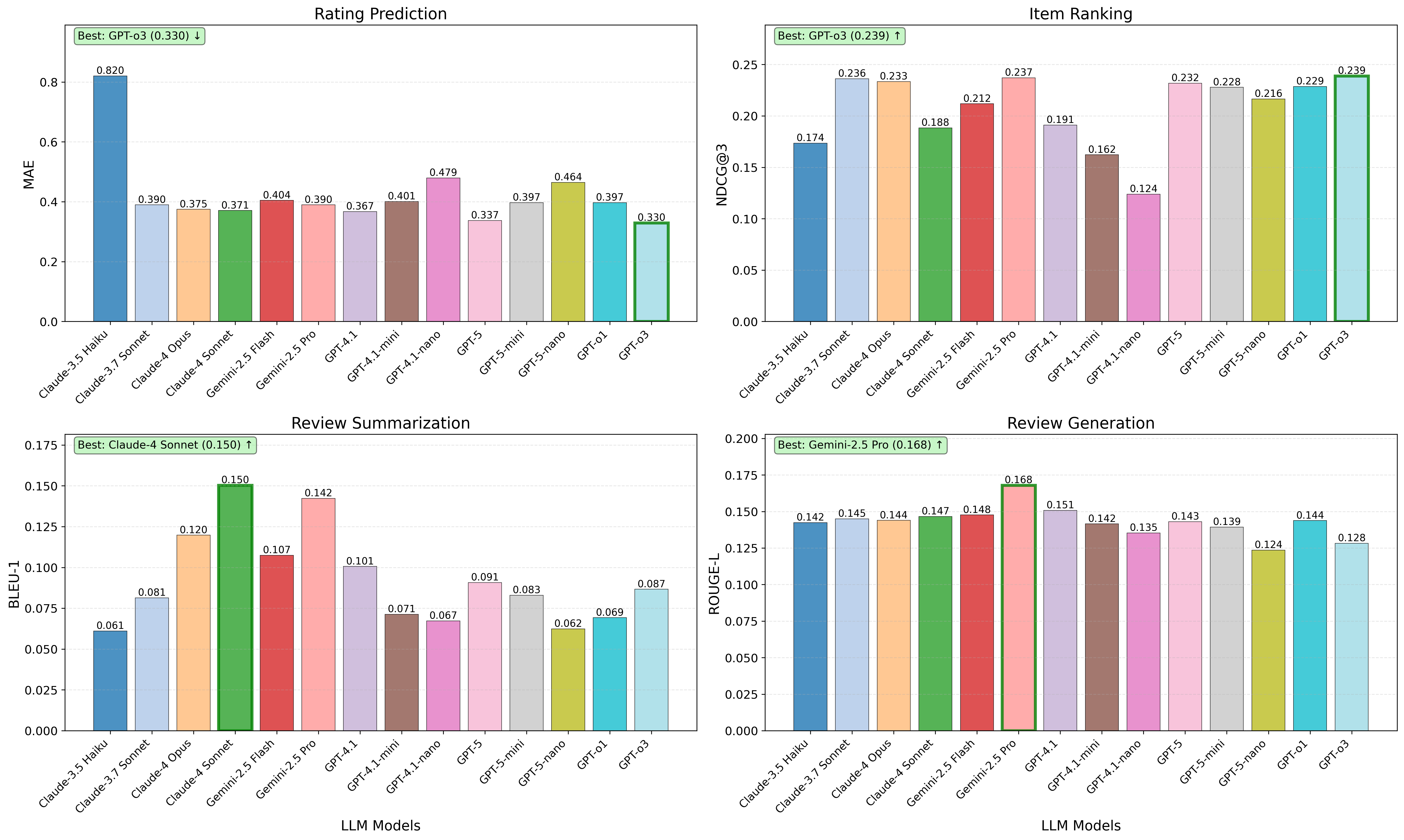}
    \caption{
    Performance comparisons of 14 frontier long-context LLMs on Books based on A-Mem.
    }
    \label{fig:sd_amem_books}
\end{figure*}

\begin{figure*}[!ht]
    \centering
    \includegraphics[width=\linewidth]{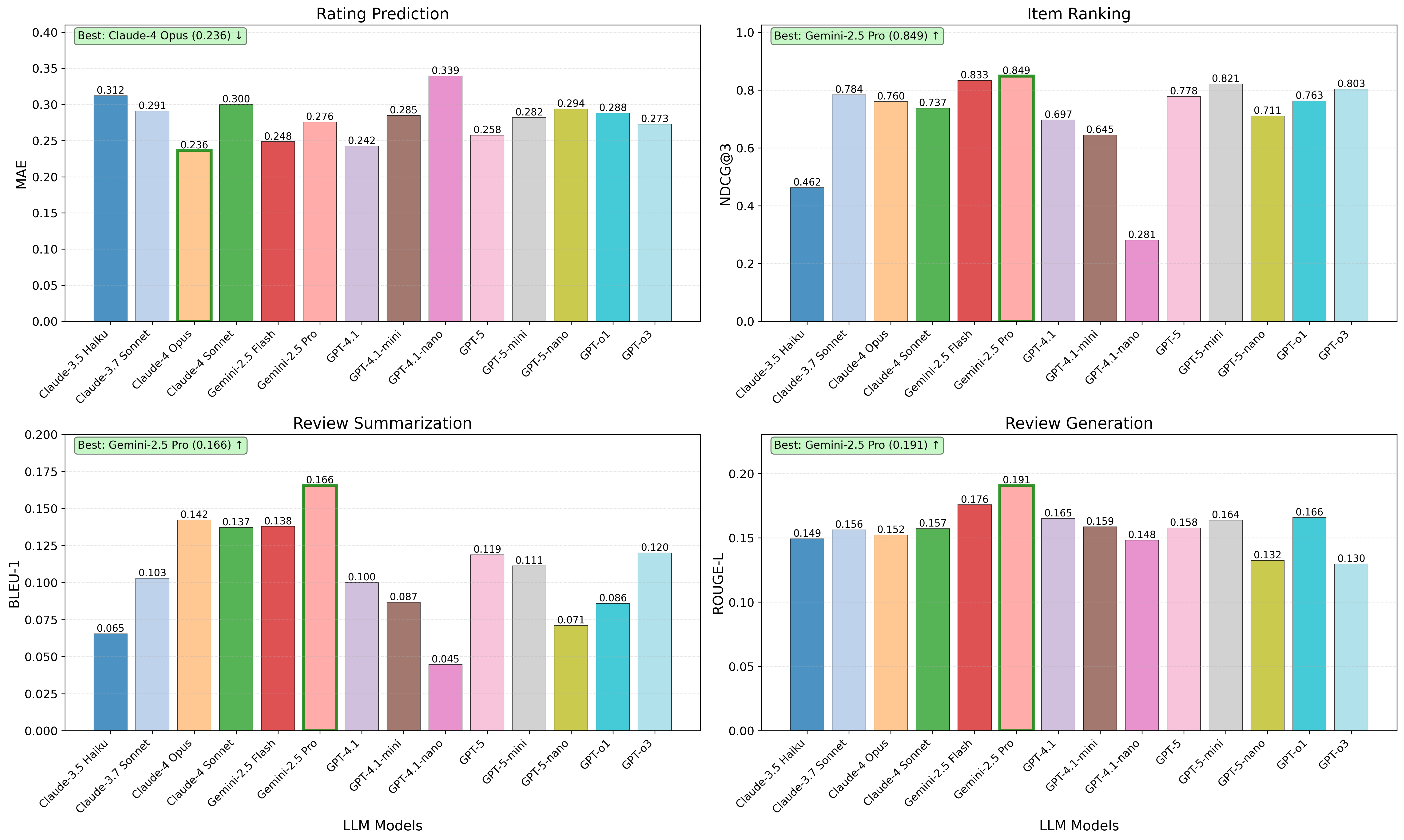}
    \caption{
    Performance comparisons of 14 frontier long-context LLMs on Home \& Kitchen based on MemoryBank.
    }
    \label{fig:sd_memorybank_home}
\end{figure*}

\begin{figure*}[!ht]
    \centering
    \includegraphics[width=\linewidth]{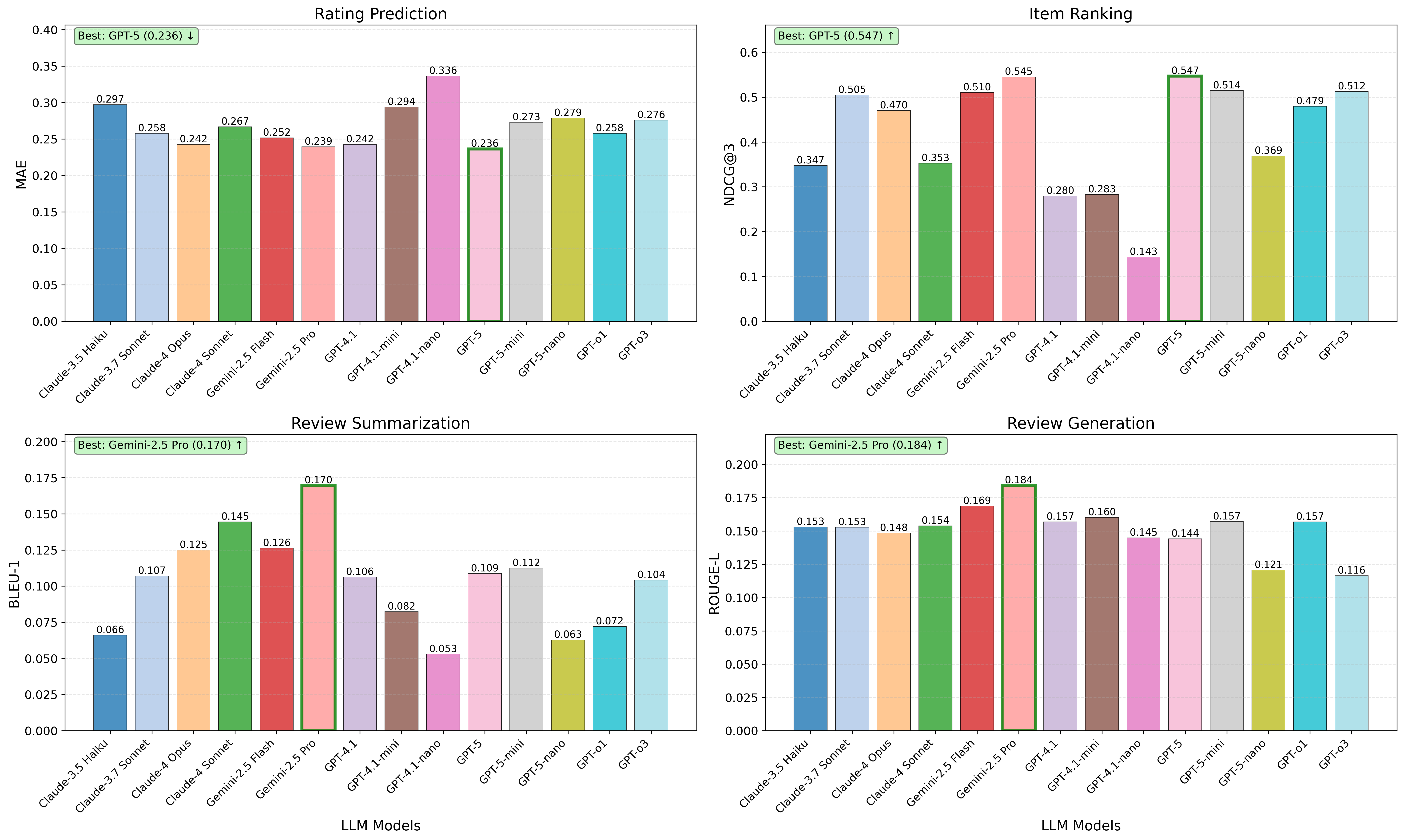}
    \caption{
    Performance comparisons of 14 frontier long-context LLMs on Home \& Kitchen based on A-Mem.
    }
    \label{fig:sd_amem_home}
\end{figure*}

Figure~\ref{fig:sd_memorybank_books}, Figure~\ref{fig:sd_amem_books}, Figure~\ref{fig:sd_memorybank_home}, and Figure~\ref{fig:sd_amem_home} compare 14 frontier long-context LLMs on Books and Home \& Kitchen under two representative memory methods (MemoryBank and A-Mem). Across both domains, we observe a stable {task-wise} pattern: (i) performance varies substantially across task types and models that excel at decision-oriented tasks (rating and ranking) are often different from those leading language generation tasks (summarization and generation); (ii) language tasks are relatively consistent across memory choices, where Gemini-family models tend to stay strong on generation and Claude-family models remain competitive on summarization, and GPT-family perform the best on decision tasks (rating/ranking); and (iii) domain difficulty differs: \textit{Home \& Kitchen} typically yields higher and more separated ranking scores than Books, suggesting stronger item-attribute regularities and cleaner preference signals in the former, whereas Books appears more ambiguous and thus compresses model gaps. 

A second key observation is that the memory method changes which backbones benefit most, especially for rating/ranking: MemoryBank tends to amplify backbones that are already strong at holistic language modeling and preference aggregation, thereby improving decision-oriented tasks such as rating prediction and item ranking. In contrast, A-Mem more frequently shifts the advantage toward models that can effectively leverage structured and explicitly linked memory notes, which proves beneficial for generation-related tasks.
Mechanically, the difference arises from how the two memory systems organize and expose user information. MemoryBank emphasizes continuous memory synthesis into a user-centric profile with retrieval for personalization. Therefore, the joint memory can support the decision-making task, such as rating and ranking. Whereas A-Mem organizes memories as structured ''notes'' with dynamic indexing and linking, enabling compositional traversal over an evolving memory graph, which better supports generative tasks.

\section{Evaluation of Different Methods in Single Domain Settings}
\label{app:methods_single_domain}

Table~\ref{tab:sd_method_comparison_electronics},  Table~\ref{tab:sd_method_comparison_home_kitchen}, and Table~\ref{tab:sd_method_comparison_personal_care} are jointly show that memory methods improve personalization in a strongly task- and domain-dependent way, with no universally dominant design. Across \textit{Electronics}, \textit{Home \& Kitchen}, and \textit{Personal Care}, retrieval and selection-style methods (e.g., Mem0/LoCoMo) most consistently boost ranking (often producing large NDCG gains over long-context prompting). In contrast, structured and profile-style memories (e.g., MemoryBank, A-Mem) more often yield balanced improvements. These highlight that cross-task gains depend not only on memory content but also on how each backbone consumes retrieved and condensed evidence. 

On \textit{Home \& Kitchen}, ranking benefits the most from retrieval/selection, while rating and generation favor more structured memories and backbone strengths.
Compared to long-context prompting, Mem0, LoCoMo, and MemoryBank substantially increase ranking scores for GPT-5 and Claude-4 Sonnet, whereas A-Mem is more competitive on rating (e.g., lowest MAE for GPT-5) but does not match the best ranking variants, suggesting a trade-off between decision-time evidence selection and preference calibration via structured memory. 
For generation tasks, Gemini-2.5 Pro remains the strongest overall generator in this domain and shows its best generation under MemoryBank-like profile memory, while GPT-5 stays comparatively stronger on decision metrics when paired with the right memory routing.

Overall, for generation tasks, Gemini-2.5 Pro remains the strongest generator in this domain, achieving its best performance when paired with MemoryBank-style profile memories, while GPT-5 continues to dominate decision-oriented metrics when combined with appropriate memory strategies.

\begin{table*}[t]
\centering
\footnotesize
\setlength{\tabcolsep}{0.8pt}
\small
\renewcommand{\arraystretch}{1.15}

\begin{tabularx}{\textwidth}{C{1.9cm} L{2.2cm} C{1cm} C{1.1cm} C{1.1cm} C{1.1cm} C{1.5cm} C{1.2cm} C{1.5cm} C{1.2cm}}
\toprule
\textbf{Method} & \textbf{Model} &
\multicolumn{2}{c}{\textbf{Rating} $\downarrow$} &
\multicolumn{2}{c}{\textbf{Ranking} $\uparrow$} &
\multicolumn{2}{c}{\textbf{Summarization} $\uparrow$} &
\multicolumn{2}{c}{\textbf{Generation} $\uparrow$} \\

& &
\textbf{MAE} & \textbf{RMSE} &
\textbf{N@1} & \textbf{N@3} &
\textbf{ROUGE-L} & \textbf{BLEU-1} &
\textbf{ROUGE-L} & \textbf{BLEU-1} \\
\midrule

\rowcolor{longbg}
\cellcolor{white}{\multirow{3}{*}{\textbf{Long-Context}}}
& GPT-5
& \textbf{0.289} & \textbf{0.602}
& \textbf{0.443} & \textbf{0.604}
& 0.182 & 0.082
& 0.123 & 0.145 \\

& Claude-4 Sonnet
& 0.374 & 0.749
& 0.354 & 0.511
& \textbf{0.228} & 0.139
& 0.144 & 0.176 \\

\rowcolor{longbg}
\cellcolor{white}{}
& Gemini-2.5 Pro
& 0.337 & 0.719
& 0.378 & 0.561
& 0.222 & \textbf{0.141}
& \textbf{0.161} & \textbf{0.200} \\

\midrule

\rowcolor{memzerobg}
\cellcolor{white}{\multirow{3}{*}{\textbf{Mem0}}}
& GPT-5
& \textbf{0.354} & \textbf{0.735}
& \textbf{0.646} & \textbf{0.744}
& 0.209 & 0.115
& 0.133 & 0.153 \\

& Claude-4 Sonnet
& 0.411 & 0.741
& 0.565 & 0.661
& \textbf{0.238} & \textbf{0.145}
& 0.144 & 0.174 \\

\rowcolor{memzerobg}
\cellcolor{white}{}
& Gemini-2.5 Pro
& 0.382 & 0.760
& 0.618 & 0.718
& 0.223 & 0.135
& \textbf{0.160} & \textbf{0.192} \\

\midrule

\rowcolor{locomobg}
\cellcolor{white}{\multirow{3}{*}{\textbf{LoCoMo}}}
& GPT-5
& \textbf{0.327} & 0.716
& \textbf{0.797} & \textbf{0.855}
& 0.188 & 0.098
& 0.141 & 0.171 \\

& Claude-4 Sonnet
& 0.358 & \textbf{0.636}
& 0.740 & 0.803
& 0.216 & \textbf{0.141}
& 0.149 & 0.183 \\

\rowcolor{locomobg}
\cellcolor{white}{}
& Gemini-2.5 Pro
& 0.430 & 0.839
& 0.758 & 0.824
& \textbf{0.219} & 0.140
& \textbf{0.167} & \textbf{0.203} \\

\midrule

\rowcolor{readbg}
\cellcolor{white}{\multirow{3}{*}{\textbf{ReadAgent}}}
& GPT-5
& \textbf{0.378} & 0.735
& \textbf{0.545} & \textbf{0.662}
& 0.203 & 0.102
& 0.132 & 0.157 \\

& Claude-4 Sonnet
& 0.390 & \textbf{0.704}
& 0.472 & 0.594
& 0.216 & 0.128
& 0.142 & 0.173 \\

\rowcolor{readbg}
\cellcolor{white}{}
& Gemini-2.5 Pro
& 0.419 & 0.794
& 0.520 & 0.645
& \textbf{0.224} & \textbf{0.131}
& \textbf{0.158} & \textbf{0.189} \\

\midrule

\rowcolor{bankbg}
\cellcolor{white}{\multirow{3}{*}{\textbf{MemoryBank}}}
& GPT-5
& \textbf{0.341} & 0.716
& 0.663 & 0.763
& 0.201 & 0.114
& 0.143 & 0.172 \\

& Claude-4 Sonnet
& 0.362 & \textbf{0.660}
& 0.593 & 0.684
& 0.232 & \textbf{0.153}
& 0.152 & 0.183 \\

\rowcolor{bankbg}
\cellcolor{white}{}
& Gemini-2.5 Pro
& 0.415 & 0.727
& \textbf{0.703} & \textbf{0.801}
& \textbf{0.236} & 0.143
& \textbf{0.170} & \textbf{0.202} \\

\midrule

\rowcolor{amembg}
\cellcolor{white}{\multirow{3}{*}{\textbf{A-Mem}}}
& GPT-5
& 0.325 & 0.681
& \textbf{0.504} & \textbf{0.650}
& 0.198 & 0.105
& 0.132 & 0.153 \\

& Claude-4 Sonnet
& \textbf{0.321} & \textbf{0.628}
& 0.435 & 0.560
& \textbf{0.246} & \textbf{0.155}
& 0.149 & 0.183 \\

\rowcolor{amembg}
\cellcolor{white}{}
& Gemini-2.5 Pro
& 0.362 & 0.762
& 0.492 & 0.643
& 0.235 & \textbf{0.155}
& \textbf{0.166} & \textbf{0.202} \\

\bottomrule
\end{tabularx}

\caption{Comparison of memory methods on \textit{Electronics} using representative LLMs.}
\label{tab:sd_method_comparison_electronics}
\end{table*}

\begin{table*}[t]
\centering
\footnotesize
\setlength{\tabcolsep}{0.8pt}
\small
\renewcommand{\arraystretch}{1.15}

\begin{tabularx}{\textwidth}{C{1.9cm} L{2.2cm} C{1cm} C{1.1cm} C{1.1cm} C{1.1cm} C{1.5cm} C{1.2cm} C{1.5cm} C{1.2cm}}
\toprule
\textbf{Method} & \textbf{Model} &
\multicolumn{2}{c}{\textbf{Rating} $\downarrow$} &
\multicolumn{2}{c}{\textbf{Ranking} $\uparrow$} &
\multicolumn{2}{c}{\textbf{Summarization} $\uparrow$} &
\multicolumn{2}{c}{\textbf{Generation} $\uparrow$} \\

& &
\textbf{MAE} & \textbf{RMSE} &
\textbf{N@1} & \textbf{N@3} &
\textbf{ROUGE-L} & \textbf{BLEU-1} &
\textbf{ROUGE-L} & \textbf{BLEU-1} \\
\midrule

\rowcolor{longbg}
\cellcolor{white}{\multirow{3}{*}{\textbf{Long-Context}}}
& GPT-5
& \textbf{0.261} & \textbf{0.551}
& \textbf{0.446} & \textbf{0.610}
& 0.185 & 0.077
& 0.134 & 0.140 \\

& Claude-4 Sonnet
& 0.330 & 0.630
& 0.276 & 0.400
& 0.189 & 0.117
& 0.145 & 0.158 \\

\rowcolor{longbg}
\cellcolor{white}{}
& Gemini-2.5 Pro
& 0.303 & 0.679
& 0.315 & 0.444
& \textbf{0.222} & \textbf{0.133}
& \textbf{0.178} & \textbf{0.207} \\

\midrule

\rowcolor{memzerobg}
\cellcolor{white}{\multirow{3}{*}{\textbf{Mem0}}}
& GPT-5
& \textbf{0.261} & \textbf{0.561}
& 0.606 & 0.707
& 0.221 & 0.122
& 0.151 & 0.163 \\

& Claude-4 Sonnet
& 0.339 & 0.623
& 0.521 & 0.613
& 0.227 & 0.152
& 0.152 & 0.167 \\

\rowcolor{memzerobg}
\cellcolor{white}{}
& Gemini-2.5 Pro
& 0.306 & 0.654
& \textbf{0.652} & \textbf{0.746}
& \textbf{0.231} & \textbf{0.155}
& \textbf{0.184} & \textbf{0.215} \\

\midrule

\rowcolor{locomobg}
\cellcolor{white}{\multirow{3}{*}{\textbf{LoCoMo}}}
& GPT-5
& \textbf{0.268} & \textbf{0.558}
& \textbf{0.824} & \textbf{0.889}
& 0.219 & 0.121
& 0.157 & 0.176 \\

& Claude-4 Sonnet
& 0.324 & 0.583
& 0.755 & 0.839
& 0.204 & 0.126
& 0.151 & 0.169 \\

\rowcolor{locomobg}
\cellcolor{white}{}
& Gemini-2.5 Pro
& 0.310 & 0.632
& 0.798 & 0.855
& \textbf{0.241} & \textbf{0.164}
& \textbf{0.191} & \textbf{0.215} \\

\midrule

\rowcolor{readbg}
\cellcolor{white}{\multirow{3}{*}{\textbf{ReadAgent}}}
& GPT-5
& \textbf{0.255} & \textbf{0.534}
& 0.409 & \textbf{0.557}
& 0.184 & 0.091
& 0.149 & 0.163 \\

& Claude-4 Sonnet
& 0.345 & 0.661
& 0.412 & 0.547
& 0.193 & 0.117
& 0.152 & 0.171 \\

\rowcolor{readbg}
\cellcolor{white}{}
& Gemini-2.5 Pro
& 0.339 & 0.726
& \textbf{0.430} & 0.573
& \textbf{0.215} & \textbf{0.132}
& \textbf{0.180} & \textbf{0.211} \\

\midrule

\rowcolor{bankbg}
\cellcolor{white}{\multirow{3}{*}{\textbf{MemoryBank}}}
& GPT-5
& \textbf{0.258} & \textbf{0.564}
& 0.688 & 0.778
& 0.210 & 0.119
& 0.158 & 0.177 \\

& Claude-4 Sonnet
& 0.300 & 0.564
& 0.645 & 0.737
& 0.219 & 0.137
& 0.157 & 0.176 \\

\rowcolor{bankbg}
\cellcolor{white}{}
& Gemini-2.5 Pro
& 0.276 & 0.575
& \textbf{0.764} & \textbf{0.849}
& \textbf{0.235} & \textbf{0.166}
& \textbf{0.191} & \textbf{0.219} \\

\midrule

\rowcolor{amembg}
\cellcolor{white}{\multirow{3}{*}{\textbf{A-Mem}}}
& GPT-5
& \textbf{0.236} & \textbf{0.505}
& \textbf{0.412} & \textbf{0.547}
& 0.209 & 0.109
& 0.144 & 0.153 \\

& Claude-4 Sonnet
& 0.267 & 0.539
& 0.239 & 0.353
& 0.220 & 0.145
& 0.154 & 0.168 \\

\rowcolor{amembg}
\cellcolor{white}{}
& Gemini-2.5 Pro
& 0.239 & 0.531
& 0.403 & 0.545
& \textbf{0.249} & \textbf{0.170}
& \textbf{0.184} & \textbf{0.214} \\

\bottomrule
\end{tabularx}

\caption{Comparison of memory methods on \textit{Home \& Kitchen} using representative LLMs.}
\label{tab:sd_method_comparison_home_kitchen}
\end{table*}

\begin{table*}[t]
\centering
\footnotesize
\setlength{\tabcolsep}{0.8pt}
\small
\renewcommand{\arraystretch}{1.15}

\begin{tabularx}{\textwidth}{C{1.9cm} L{2.2cm} C{1cm} C{1.1cm} C{1.1cm} C{1.1cm} C{1.5cm} C{1.2cm} C{1.5cm} C{1.2cm}}
\toprule
\textbf{Method} & \textbf{Model} &
\multicolumn{2}{c}{\textbf{Rating} $\downarrow$} &
\multicolumn{2}{c}{\textbf{Ranking} $\uparrow$} &
\multicolumn{2}{c}{\textbf{Summarization} $\uparrow$} &
\multicolumn{2}{c}{\textbf{Generation} $\uparrow$} \\

& &
\textbf{MAE} & \textbf{RMSE} &
\textbf{N@1} & \textbf{N@3} &
\textbf{ROUGE-L} & \textbf{BLEU-1} &
\textbf{ROUGE-L} & \textbf{BLEU-1} \\
\midrule

\rowcolor{longbg}
\cellcolor{white}{\multirow{3}{*}{\textbf{Long-Context}}}
& GPT-5
& \textbf{0.279} & \textbf{0.597}
& \textbf{0.315} & \textbf{0.435}
& 0.161 & 0.053
& 0.131 & 0.139 \\

& Claude-4 Sonnet
& 0.432 & 0.767
& 0.203 & 0.304
& 0.196 & \textbf{0.118}
& 0.138 & 0.148 \\

\rowcolor{longbg}
\cellcolor{white}{}
& Gemini-2.5 Pro
& 0.323 & 0.665
& 0.266 & 0.417
& \textbf{0.200} & 0.110
& \textbf{0.183} & \textbf{0.204} \\

\midrule

\rowcolor{memzerobg}
\cellcolor{white}{\multirow{3}{*}{\textbf{Mem0}}}
& GPT-5
& \textbf{0.331} & \textbf{0.635}
& 0.578 & 0.685
& 0.175 & 0.082
& 0.150 & 0.164 \\

& Claude-4 Sonnet
& 0.361 & 0.672
& 0.583 & 0.683
& 0.197 & 0.111
& 0.151 & 0.163 \\

\rowcolor{memzerobg}
\cellcolor{white}{}
& Gemini-2.5 Pro
& 0.353 & 0.704
& \textbf{0.612} & \textbf{0.725}
& \textbf{0.214} & \textbf{0.131}
& \textbf{0.182} & \textbf{0.201} \\

\midrule

\rowcolor{locomobg}
\cellcolor{white}{\multirow{3}{*}{\textbf{LoCoMo}}}
& GPT-5
& \textbf{0.315} & \textbf{0.633}
& \textbf{0.823} & \textbf{0.897}
& 0.192 & 0.095
& 0.160 & 0.177 \\

& Claude-4 Sonnet
& 0.362 & 0.635
& 0.802 & 0.885
& 0.198 & 0.115
& 0.158 & 0.172 \\

\rowcolor{locomobg}
\cellcolor{white}{}
& Gemini-2.5 Pro
& 0.375 & 0.670
& 0.814 & 0.889
& \textbf{0.208} & \textbf{0.133}
& \textbf{0.189} & \textbf{0.211} \\

\midrule

\rowcolor{readbg}
\cellcolor{white}{\multirow{3}{*}{\textbf{ReadAgent}}}
& GPT-5
& \textbf{0.339} & \textbf{0.633}
& 0.404 & 0.535
& 0.170 & 0.069
& 0.148 & 0.164 \\

& Claude-4 Sonnet
& 0.427 & 0.703
& 0.380 & 0.525
& \textbf{0.208} & 0.107
& 0.149 & 0.164 \\

\rowcolor{readbg}
\cellcolor{white}{}
& Gemini-2.5 Pro
& 0.365 & 0.688
& \textbf{0.417} & \textbf{0.567}
& 0.206 & \textbf{0.113}
& \textbf{0.172} & \textbf{0.195} \\

\midrule

\rowcolor{bankbg}
\cellcolor{white}{\multirow{3}{*}{\textbf{MemoryBank}}}
& GPT-5
& \textbf{0.276} & \textbf{0.582}
& 0.742 & 0.823
& 0.195 & 0.095
& 0.163 & 0.175 \\

& Claude-4 Sonnet
& 0.318 & 0.629
& 0.664 & 0.771
& 0.219 & 0.135
& 0.161 & 0.176 \\

\rowcolor{bankbg}
\cellcolor{white}{}
& Gemini-2.5 Pro
& 0.349 & 0.688
& \textbf{0.768} & \textbf{0.859}
& \textbf{0.221} & \textbf{0.138}
& \textbf{0.192} & \textbf{0.213} \\

\midrule

\rowcolor{amembg}
\cellcolor{white}{\multirow{3}{*}{\textbf{A-Mem}}}
& GPT-5
& \textbf{0.299} & \textbf{0.589}
& \textbf{0.365} & \textbf{0.496}
& 0.186 & 0.082
& 0.145 & 0.156 \\

& Claude-4 Sonnet
& 0.341 & 0.627
& 0.336 & 0.466
& \textbf{0.230} & \textbf{0.142}
& 0.157 & 0.170 \\

\rowcolor{amembg}
\cellcolor{white}{}
& Gemini-2.5 Pro
& 0.305 & 0.619
& 0.339 & 0.477
& 0.218 & 0.128
& \textbf{0.188} & \textbf{0.213} \\

\bottomrule
\end{tabularx}

\caption{Comparison of memory methods on \textit{Personal Care} using representative LLMs.}
\label{tab:sd_method_comparison_personal_care}
\end{table*}

\section{Prompt Templates for Personalized Tasks}
\label{sec:prompt-templates}

We present the exact prompt templates used for all four personalized tasks: rating prediction, item ranking, review summarization, and review generation. Each prompt follows a unified structure that conditions on the user's recent interaction history (serving as an implicit memory of preferences and writing style) and task-specific inputs (e.g., target product, candidate set, or review text). We enforce strict output formats to ensure consistency and enable reliable automatic evaluation across tasks.

\vspace{0.5em}

\begin{tcolorbox}[breakable, title=Rating Prediction Prompt, colback=gray!5, colframe=gray!50]
\small
\ttfamily
You are a personalized rating prediction system. Based on a user's purchase and review history, predict what rating (1-5) they would give to a new product.

\vspace{2mm}
User's Relevant Review History:

1. [Formatted Interaction 1] \\
2. [Formatted Interaction 2] \\
\ldots \\
n. [Formatted Interaction n]

\vspace{2mm}
Target Product to Rate: \\
{}[Formatted Target Item without Rating and Review Text]

\vspace{2mm}
Based on this user's history and preferences, predict the rating (1-5) they would give to the target product.

Respond with ONLY a single integer between 1 and 5. Output only the number.
\end{tcolorbox}

\vspace{0.5em}

\begin{tcolorbox}[breakable, title=Item Ranking Prompt, colback=gray!5, colframe=gray!50]
\small
\ttfamily
You are a personalized item recommendation system. Based on a user's purchase and review history, rank the candidate items in order of relevance (most likely to purchase first).

\vspace{2mm}
User's Relevant Purchase and Review History:

1. [Formatted Interaction 1] \\
2. [Formatted Interaction 2] \\
\ldots \\
n. [Formatted Interaction n]

\vspace{2mm}
Candidate Items to Rank:

1. [Candidate Item 1] \\
2. [Candidate Item 2] \\
\ldots \\
k. [Candidate Item k]

\vspace{2mm}
Based on this user's history and preferences, rank the candidate items from most to least relevant.

Respond with ONLY the ranked list of item indices (e.g., 3, 1, 2, ...).
\end{tcolorbox}

\vspace{0.5em}

\begin{tcolorbox}[breakable, title=Review Summarization Prompt, colback=gray!5, colframe=gray!50]
\small
\ttfamily
You are a personalized review summarization system. Based on a user's review history and writing style, generate a concise review title that summarizes their full review.

\vspace{2mm}
User's Relevant Review History (showing their writing style and preferences):

1. [Formatted Interaction 1] \\
2. [Formatted Interaction 2] \\
\ldots \\
n. [Formatted Interaction n]

\vspace{2mm}
Target Product and Review to Summarize: \\
{}[Formatted Target Item without Rating and Review Text] \\
User's Rating: [Target Rating]/5 \\
User's Full Review: [Target Review Text]

\vspace{2mm}
Based on this user's writing style and the full review, generate a concise review title.

The title should capture the main sentiment and key points in the user's style.

Respond with ONLY the review title (3--10 words).
\end{tcolorbox}

\vspace{0.5em}

\begin{tcolorbox}[breakable, title=Review Generation Prompt, colback=gray!5, colframe=gray!50]
\small
\ttfamily
You are a personalized review generation system. Based on a user's review history and writing style, generate a detailed review text for a product they rated.

\vspace{2mm}
User's Relevant Review History (showing their writing style and preferences):

1. [Formatted Interaction 1] \\
2. [Formatted Interaction 2] \\
\ldots \\
n. [Formatted Interaction n]

\vspace{2mm}
Target Product to Review: \\
{}[Formatted Target Item without Rating and Review Text] \\
User's Rating: [Target Rating]/5

\vspace{2mm}
Based on this user's writing style, preferences, and the rating they gave, generate a detailed review text.

The review should reflect the user's typical style, length, and sentiment.

Respond with ONLY the review text.
\end{tcolorbox}

\end{document}